\documentclass[10pt,twocolumn,letterpaper]{article}

\usepackage{cvpr}
\usepackage{times}
\usepackage{epsfig}
\usepackage{graphicx}
\usepackage{amsmath}
\usepackage{amssymb}
\usepackage{algorithm}
\usepackage{algorithmic}
\usepackage{mathrsfs}
\usepackage{latexsym}
\usepackage{booktabs}
\usepackage{multirow}
\usepackage{underscore}
\usepackage{lineno}
\usepackage{color}
\usepackage{lipsum}
\usepackage{setspace}
\usepackage{lipsum}
\usepackage{subfig}
\usepackage[noblocks]{authblk}
\usepackage{setspace}
\usepackage{url}
\usepackage{xcolor}
\definecolor{newcolor}{rgb}{.8,.349,.1}

\makeatletter
\renewcommand\normalsize{%
   \@setfontsize\normalsize\@xpt\@xiipt
   \abovedisplayskip 5\p@ \@plus2\p@ \@minus5\p@
   \abovedisplayshortskip \z@ \@plus3\p@
   \belowdisplayshortskip 7\p@ \@plus3\p@ \@minus1\p@
   \belowdisplayskip \abovedisplayskip
   \let\@listi\@listI}
\makeatother

\usepackage[pagebackref=true,breaklinks=true,letterpaper=true,colorlinks,citecolor=blue,linkcolor=blue,bookmarks=false]{hyperref}

 \cvprfinalcopy 

\def\cvprPaperID{1353} 

\ifcvprfinal\fi
\begin{document}

\title{Learning Spatial-Temporal Regularized Correlation Filters for Visual Tracking}

\author{\large Feng Li}
\author{\large Cheng Tian}
\author{\large Wangmeng Zuo \thanks{Corresponding author.}}
\affil{\normalsize School of Computer Science and Technology, Harbin Institute of Technology, China}
\author{\large Lei Zhang}
\affil{\normalsize Department of Computing, The Hong Kong Polytechnic University, China}
\author{\large Ming-Hsuan Yang}
\affil{\normalsize School of Engineering, University of California, Merced, USA
\authorcr{\tt\small{fengli\_hit@hotmail.com, tcoperator@163.com, wmzuo@hit.edu.cn, csdzhang@comp.polyu.edu.hk, mhyang@ucmerced.edu}}}


\maketitle

\newcommand\blfootnote[1]{%
\begingroup
\renewcommand\thefootnote{}\footnote{#1}%
\addtocounter{footnote}{-1}%
\endgroup
}

\blfootnote{This work is supported by NSFC grants (No.s 61671182 and 61471146) and HK RGC GRF grant (PolyU 152240/15E).}

\begin{abstract}
Discriminative Correlation Filters (DCF) are efficient in visual tracking but suffer from unwanted boundary effects.
Spatially Regularized DCF (SRDCF) has been suggested to resolve this issue by enforcing spatial penalty on DCF coefficients, which, inevitably, improves the tracking performance at the price of increasing complexity.
To tackle online updating, SRDCF formulates its model on multiple training images, further adding difficulties in improving efficiency.
In this work, by introducing temporal regularization to SRDCF with single sample, we present our spatial-temporal regularized correlation filters (STRCF).
%
The STRCF formulation can not only serve as a reasonable approximation to SRDCF with multiple training samples, but also provide a more robust appearance model than SRDCF in the case of large appearance variations.
Besides, it can be efficiently solved via the alternating direction method of multipliers (ADMM).
By incorporating both temporal and spatial regularization, our STRCF can handle boundary effects without much loss in efficiency and achieve superior performance over SRDCF in terms of accuracy and speed.
%
%
Compared with SRDCF, STRCF with hand-crafted features provides a $5\times$ speedup and achieves a gain of 5.4\% and 3.6\% AUC score on OTB-2015 and Temple-Color, respectively.
Moreover, {STRCF with deep features also performs favorably against state-of-the-art trackers and achieves an AUC score of 68.3\% on OTB-2015.}

\end{abstract}

\vspace{-6mm}
\section{Introduction}\label{sec:Intro}

\begin{figure}[!htbp]
\setlength{\abovecaptionskip}{0.1cm}
\setlength{\belowcaptionskip}{-0.6cm}
\centering
\subfloat[]{\label{fig:1(a)}
  \includegraphics[width=0.48\textwidth]{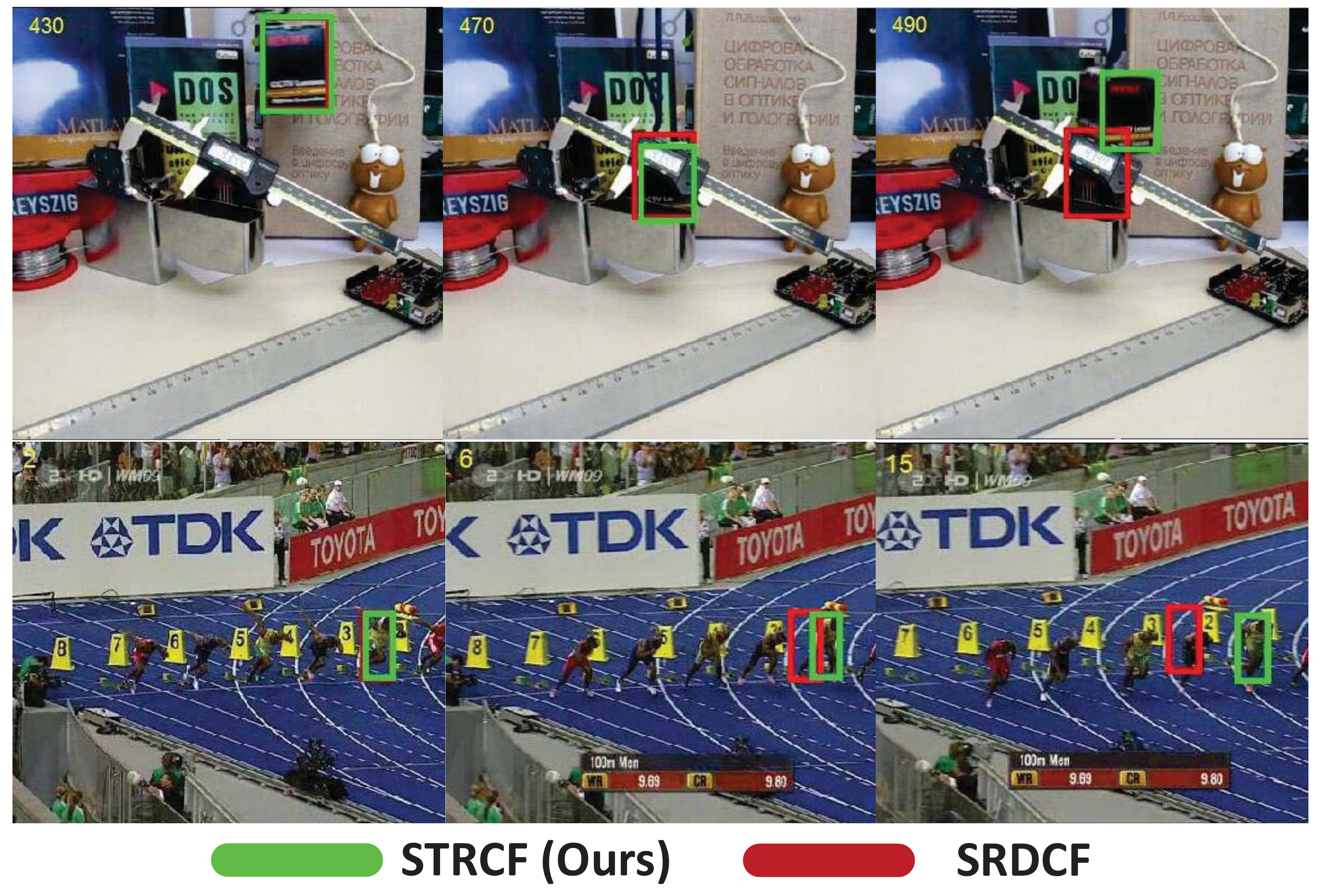}}\
\subfloat[]{\label{fig:1(b)}
  \includegraphics[width=0.48\textwidth]{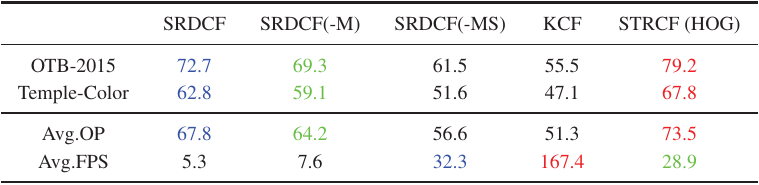}}
\caption{\small{(a) The results of STRCF and SRDCF \cite{danelljan2015learning} on two sequences with occlusion and deformation. (b) A comparison of SRDCF variants and STRCF using HOG feature in terms of mean OP (\%) and speed (FPS) on OTB-2015 and Temple-Color. The best three results are shown in {\color{red}red}, {\color{blue}blue} and {\color{green}green} fonts, respectively.}}
\end{figure}

Recent years have witnessed the rapid advances of discriminative correlation filters (DCFs) in visual tracking.
Benefited from the periodic assumption of training samples, the DCF can be learned very efficiently in the frequency domain via fast Fourier transform (FFT).
For example, the tracking speed of the earliest DCF-based tracker, \ie, MOSSE~\cite{bolme2010visual}, can reach 700 frames per second (FPS).
Along with the introduction of feature representation~\cite{Danelljan2014Adaptive,ma2015hierarchical}, nonlinear kernel~\cite{henriques2015high}, scale estimation~\cite{danelljan2016discriminative,Li2017Integrating,li2014scale}, max-margin classifiers~\cite{zuo2016learning}, spatial regularization~\cite{danelljan2015learning,GaloogahiSL14}, and continuous convolution~\cite{Danelljan2016CCOT}, DCF-based trackers have been greatly improved and significantly advanced the state-of-the-art tracking accuracy.
However, such performance improvement is not obtained without any extra cost.
Most top-ranked trackers, \eg, SRDCF~\cite{danelljan2015learning} and C-COT~\cite{Danelljan2016CCOT}, have gradually lost the characteristic speed and realtime capability of early DCF-based trackers.
For example, the speed of SRDCF~\cite{danelljan2015learning} using the hand-crafted HOG feature is $\sim$6 FPS, while that of the baseline KCF~\cite{henriques2015high} is $\sim$170 FPS.

For better understanding on this issue, we dissect the tradeoff between accuracy and speed in SRDCF.
In general, the inefficiency of SRDCF can be attributed to three factors: (i) scale estimation, (ii) spatial regularization, and (iii) formulation on large training set.
Fig.~\ref{fig:1(b)} lists the tracking speed and accuracy of SRDCF and its variants on two popular benchmarks, including SRDCF($-$M) (\ie, removing (iii)), SRDCF($-$MS) (\ie, removing (ii)\&(iii)), and KCF (\ie, removing (i)\&(ii)\&(iii)).
We note that when removing (iii), linear interpolation~\cite{bolme2010visual,danelljan2016discriminative} is adopted as an alternative strategy for online model updating.
From Fig.~\ref{fig:1(b)}, it can be seen that the tracker still maintains its real-time ability ($\sim 33 FPS$) when adding scale estimation.
But the tracking speed decreases significantly with the further introduction of spatial regularization and formulation on large training set.
Therefore, it is valuable to develop a solution for taking use of (ii) and (iii) without much loss in efficiency.

In this paper, we study the solution for taking the benefit of spatial regularization and formulation on large training set without much loss in efficiency.
On the one hand, the high complexity of SRDCF mainly comes at the formulation on multiple training images.
By removing the constraint, SRDCF with single image can be efficiently solved via ADMM.
Due to the convexity of SRDCF, the ADMM can also guarantee to converge to global optimum.
On the other hand, in SRDCF spatial regularization is integrated into the formulation on multiple training images for the coupling of DCF learning and model updating, which does benefit the tracking accuracy.
Motivated by online Passive-Aggressive (PA) learning~\cite{Crammer2006Online}, we introduce a temporal regularization to SRDCF with single image, resulting in our spatial-temporal regularized correlation filters (STRCF).
STRCF is a rational approximation of the full SRDCF formulation on multiple training images, and can also be exploited for simultaneous DCF learning and model updating. Besides, the ADMM algorithm can also be directly used to solve STRCF.
Thus, our STRCF incorporates both spatial and temporal regularization into DCF, and can be adopted to speed up SRDCF.

Furthermore, as an extension of online PA algorithm~\cite{Crammer2006Online}, STRCF can also provide a more robust appearance model than SRDCF in the case of significant appearance variations.
Fig.~\ref{fig:1(a)} illustrates the tracking results on two sequences with occlusion and deformation.
Compared with SRDCF, we can see that, with the introduction of the temporal regularization, STRCF performs more robustly to occlusion while adapting well to large appearance variation.
From Fig.~\ref{fig:1(b)}, STRCF not only runs at real-time tracking speed ($\sim 30 FPS$), but also
leads to $+5.7\%$ performance gain over SRDCF by average mean OP on two datasets.
To sum up, STRCF can achieve remarkable improvements over the baseline SRDCF on all the datasets, and runs at more than $5\times$ faster tracking speed.

We perform comparative experiments on several benchmarks, including OTB-2015~\cite{wu2015object}, Temple-Color~\cite{Liang2015Encoding}, and VOT-2016~\cite{kristan2016visual}.
STRCF performs favorably in terms of accuracy, robustness and speed in comparison with the state-of-the-art CF-based and CNN trackers.

The contributions of this paper are as follows:
\begin{itemize}
  \item A STRCF model is presented by incorporating both spatial and temporal regularization into the DCF framework. Based on online PA, STRCF can not only serve as a rational approximation of the SRDCF formulation on multiple training images, but also provide a more robust appearance model than SRDCF in the case of large appearance variations.
  \item An ADMM algorithm is developed for solving STRCF efficiently, where each sub-problem has the closed-form solution. And our algorithm can empirically converge within very few iterations.
  \item {Our STRCF with hand-crafted feature can run in real-time, achieves notable improvements over SRDCF by tracking accuracy. Furthermore, our STRCF with deep features performs favorably in comparison with the state-of-the-art trackers~\cite{Danelljan2016ECO,Danelljan2016CCOT}.}
\end{itemize}
\vspace{-2mm}

\section{Related Work}

This section first provides a brief survey on DCF trackers and then focuses on spatial regularization and formulation on large training set that are most relevant to our STRCF.

\subsection{Discriminative Correlation Filters}

Using DCFs for adaptive tracking starts with MOSSE \cite{bolme2010visual}, which learns the CFs with few samples in the frequency domain.
Notable improvements have been made to this popular tracker to address several limiting issues.
For example, Henriques \etal \cite{henriques2015high} learn the kernelized CFs (KCF) via kernel trick.
The multi-channel version of MOSSE is also studied in \cite{kiani2013multi}.
And more discriminative features are widely used, such as HOG \cite{Dalal2005Histograms}, color names (CN) \cite{Danelljan2014Adaptive} and {deep CNN} features \cite{ma2015hierarchical,qi2016hedged}.
To cope with the size change and occlusion, several scale-adaptive \cite{danelljan2016discriminative,Li2017Integrating,li2014scale} and part-based trackers \cite{Liu2016SCF,Liu2015Real} are further investigated.
Besides, long-term tracking \cite{Ma2015Long}, continuous convolution \cite{Danelljan2016CCOT} and particle filter based methods \cite{Zhang2017MCPF} are also developed to improve the tracking accuracy and robustness.
Due to the space limitation, here we only review the methods from spatial regularization and formulation on large training set that are close to our algorithm.

\subsection{Spatial Regularization}

The circulant shifted samples in DCF-based trackers always suffer from periodic repetitions on boundary positions, thereby significantly degrading the tracking performance.
Several spatial regularization methods have been suggested to alleviate the unwanted boundary effects.
Galoogahi \etal \cite{GaloogahiSL14}  pre-multiply the image patches with a fixed masking matrix containing the target regions,
and then solve the constrained optimization problem via ADMM.
However, their method can only be applied to single channel DCFs.
Danelljan \etal \cite{danelljan2015learning} propose a spatial regularization term to penalize the DCF coefficients depending on their spatial locations and suggest the Gauss-Seidel algorithm to solve the resulting normal equations.
The work \cite{Cui2016Recurrently} also employs a similar spatial regularization term,
but the spatial regularization matrix is predicted with a multi-directional RNN for identifying the reliable components.
These two methods, however, are unable to exploit the circulant structure in learning, resulting in higher computational cost.
More recently, Galoogahi \etal \cite{Galoogahi2017Learning} extend \cite{GaloogahiSL14} to multiple channels and further speed up the tracker towards real-time.
Compared with these methods, our STRCF has several merits:
(1) while STRCF serves as an approximation to \cite{danelljan2015learning} on multiple training samples, it can be solved more efficiently with the proposed ADMM algorithm.
(2) with the introduction of the temporal regularization, STRCF can learn a more robust appearance model than \cite{danelljan2015learning,Galoogahi2017Learning}, thereby leading to superior tracking performance.

\subsection{Formulation on large training set}
One of the most critical challenges in visual tracking is to learn and maintain a robust and fast appearance model in the case of large appearance variations.
To this end, MOSSE \cite{bolme2010visual} implements simultaneous DCF learning and model updating by learning
the CFs with multiple training samples from historical tracking results.
Similar strategy of incorporating large training set into the formulation can also be found in \cite{Danelljan2016Adaptive,danelljan2015convolutional,Danelljan2016CCOT,kiani2013multi}.
In practice, robust CFs can be learned by taking the samples at different time instances into consideration.
However, this leads to superior performance at the price of higher computational burden.
In comparison with these methods, KCF \cite{henriques2015high} and its variants \cite{Bibi2016Target,danelljan2016discriminative} decouple the DCF learning and model updating, and further exploit the circulant structure for high efficiency.
%
%
As a result, KCF with HOG feature can run at more than 150 FPS on a single CPU.
Following this work, there also exist several heuristic methods \cite{Liu2015Real,Wang2017Large} to address the naive model updating issues.
These methods, however, obtain inferior performance than DCF-based trackers with large training set.
Compared with these trackers, STRCF can not only be solved efficiently by avoiding the deployment of large training set, but also
benefit from simultaneous DCF learning and model updating by introducing the temporal regularization.

\vspace{-2mm}

\section{Spatially Regularized DCF} \label{sec:SRDCF}

In this section, we first revisit the SRDCF tracker, and then present our STRCF model motivated by online PA.
Finally, an ADMM is developed to solve the STRCF model.
\begin{figure*}[htbp]
\vspace{-0.0in}
\setlength{\abovecaptionskip}{0.2cm}
\setlength{\belowcaptionskip}{-0.2cm}
\centering
\subfloat{%
  \includegraphics[width=0.85\textwidth]{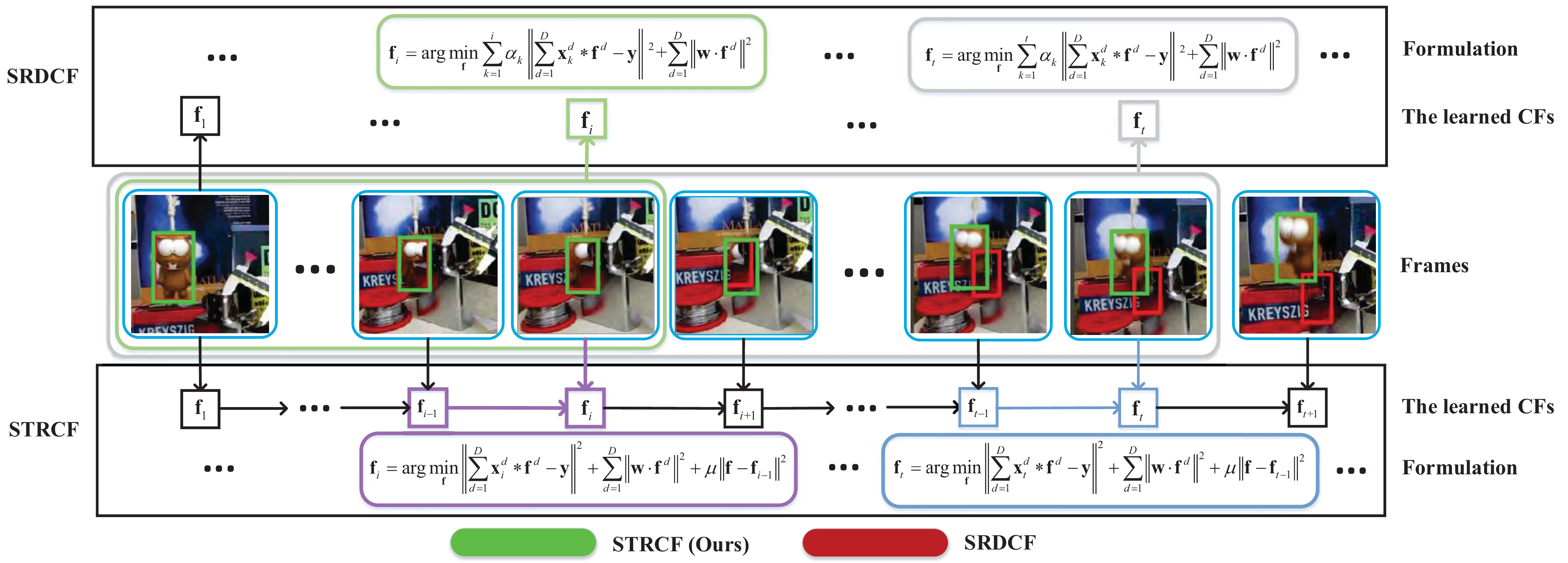}}
\caption{\small{A comparison of SRDCF and STRCF on model learning. SRDCF learns the CFs with multiple samples from historical tracking results and emphasizes more to the recent samples. Thus it may suffer from over-fitting to the recent inaccurate samples and results in tracking failure in the case of occlusion. In contrast, our STRCF trains the CF $\mathbf{f}_{{t}}$ with the sample from current frame and the learned CF $\mathbf{f}_{{t-1}}$. Benefited from online PA, STRCF can successfully follow the targets by \emph{passively} updating the CFs in the case of occlusion.}}
\label{fig:figure2}
\vspace{-0.1in}
\end{figure*}
\subsection{Revisit SRDCF}

Denote by $\mathcal{D} = \{(\mathbf{x}_k\, \mathbf{y}_k)\}_{k=1}^T$ a training set of multiple images.
Each sample $\mathbf{x}_k = [\mathbf{x}_k^1, ..., \mathbf{x}_k^D]$ consists of $D$ feature maps with size of $M \times N$.
And $\mathbf{y}_k$ is the predefined Gaussian shaped labels.
The SRDCF~\cite{danelljan2015learning} is formulated by minimizing the following objective,
%
\begin{equation}\label{equ:SRDCF}
\small
\arg\min_{\mathbf{f}} \sum_{k=1}^{T}\alpha_k \left \| \sum_{d=1}^D \mathbf{x}_k^d \ast \mathbf{f}^d - \mathbf{y}_k \right \|^2 + \sum_{d=1}^{D}\left \| \mathbf{w} \cdot \mathbf{f}^d\right \|^2,
\end{equation}
%
where $\cdot$ denotes the {Hadamard product}, $\ast$ stands for the convolution operator, $\mathbf{w}$ and $\mathbf{f}$ are the spatial regularization matrix and correlation filter, respectively.
$\alpha_k$ indicates the weight to each sample $\mathbf{x}_k$ and is set to emphasize more to the recent samples.
In~\cite{danelljan2015learning}, Danelljan \etal employ the Gauss-Seidel method to iteratively update the filters $\mathbf{f}$.
Please refer to~\cite{danelljan2015learning} for more implementation details.

However, although SRDCF is effective in suppressing the adverse boundary effects, it also increases the computational burden due to the following two reasons:

(\emph{\textbf{i}}) \emph{The failure of exploiting circulant matrix structure}. For the sake of learning a robust correlation filter $\mathbf{f}$,
DCF trackers incorporate several historical samples $\{(\mathbf{x}_k, \mathbf{y}_k)\}_{k=1}^T$ for training \cite{danelljan2016discriminative}.
However, unlike other CF-based trackers learned with only the sample from the current frame,
the formulation on multiple images breaks the circulant matrix structure, resulting in high computation burden.
As for SRDCF, the optimization becomes even more difficult due to the spatial regularization term.

(\emph{\textbf{ii}}) \emph{The large linear equations and Gauss-Seidel solver}. Eqn. (\ref{equ:SRDCF}) results in a $DMN \times DMN$ large sparse linear equation system.
While the Gauss-Seidel method is suggested to solve Eqn. (\ref{equ:SRDCF}) using the property of sparse matrix, it still remains high computational  complexity.
In addition, the SRDCF tracker also needs a long start-up time to learn the discriminative correlation filters in the first frame due to the low convergence speed of Gauss-Seidel method.

Both the spatial regularization and formulation on multiple images will break the circulant matrix structure.
Fortunately, these two issues can be circumvented to improve the tracking speed.
The formulation on multiple images can be relaxed to a STRCF model on single image by introducing the temporal regularization.
Furthermore, the introduction of spatial regularization can be addressed by exploiting an equivalent reformulation solved by ADMM efficiently.

\subsection{STRCF} \label{sec:STRCF}

In online classification, when a new instance comes on each round, the algorithm first predicts its label, and then updates the classifier based on the newly instance-label pair.
On the one hand, the learning algorithm should be \emph{passive} to make the updated classifier similar to the previous one.
On the other hand, the learning algorithm should be \emph{aggressive} to guarantee the new instance be corrected classified.
Thus, Crammer \etal~\cite{Crammer2006Online} suggest an online passive-aggressive (PA) algorithm by introducing a temporal regularization, and derive the bound on the cumulative loss of PA w.r.t. the best fixed predictor.

Motivated by PA, we introduce a temporal regularization term $\| \mathbf{f} - \mathbf{f}_{t-1} \|^2$, resulting in our spatial-temporal regularized CF (STRCF) model,
%
\begin{equation}\label{equ:STRCF}
\small
\arg\min_{\mathbf{f}} \frac{1}{2}\left \| \sum_{d=1}^D \mathbf{x}_t^d \ast \mathbf{f}^d \!-\! \mathbf{y} \right \|^2 \!+\frac{1}{2}\! \sum_{d=1}^{D}\left \| \mathbf{w} \cdot \mathbf{f}^d\right \|^2 \!+\!\frac{\mu}{2} \left \| \mathbf{f} \!-\! \mathbf{f}_{t-1} \right \|^2,
\end{equation}
%
where $\mathbf{f}_{t\!-\!1}$ denotes the CFs utilized in the $(t\!\!-\!1)$-th frame, and $\mu$ denotes the regularization parameter.
Here, $\sum_{d=1}^{D} \| \mathbf{w} \cdot \mathbf{f}^d \|^2$ denotes the spatial regularizer, and $\| \mathbf{f} \!-\! \mathbf{f}_{t-1} \|^2$ denotes the temporal regularizer.

STRCF can also be treated as an extension of online PA from two aspects:
(i) Instead of classification, STRCF is an online learning of linear regression;
(ii) Instead of instance-wise updating, the samples in STRCF come at the batch level (\ie all the shift versions of an image) on each round.
Therefore, STRCF naturally inherits the merits of online PA on adaptively balancing the tradeoff between aggressive and passive model learning, thus leading to more robust models in the case of large appearance variations.
In Fig. \ref{fig:figure2}, we compare STRCF with SRDCF on sequence \emph{Lemming} to highlight their relationships on CF model learning.
%
%
%
From it we can make the following observations:
(i) Similar to SRDCF, STRCF also implements simultaneous DCF learning and model updating with the introduction of temporal regularizer, thus can serve as a rational approximation of SRDCF with multiple training samples;
(ii) In the case of occlusion, while SRDCF suffers from over-fitting to recent corrupted samples, STRCF can alleviate this by \emph{passively} updating the CFs to keep it close to the previous ones.

\subsection{Optimization algorithm}\label{sec:optimization}

The model in Eqn. (\ref{equ:STRCF}) is convex, and can be minimized to obtain the globally optimal solution via ADMM.
{To this end, we first introduce an auxiliary variable $\mathbf{g}$ by requiring $\mathbf{f} = \mathbf{g}$ and the stepsize parameter $\gamma$,
then the Augmented Lagrangian form of Eqn. (\ref{equ:STRCF}) can be formulated as
\begin{small}
\setlength\abovedisplayskip{1pt}
\setlength\belowdisplayskip{1pt}
\begin{align}\label{equ:STRCFALM}
\small
\vspace{-0.1in}
\mathcal{L}(\mathbf{w},\mathbf{g},\mathbf{s}&)  = \frac{1}{2}\!\left \| \sum_{d=1}^D \mathbf{x}_t^d \!\ast\!\mathbf{f}^d \!-\!\mathbf{y} \right \|^2 \!+\frac{1}{2} \sum_{d=1}^{D}\left \| \mathbf{w} \!\cdot\! \mathbf{g}^d\right \|^2 \\\nonumber
&\!+\!\!\sum_{d=1}^D(\mathbf{f}^d \!\!-\! \mathbf{g}^d)^T\!\mathbf{s}^d \!\!+\!\!\frac{\gamma}{2}\! \sum_{d=1}^D\left \| \mathbf{f}^d\!\!-\! \mathbf{g}^d\right \|^2 \!\!\!+\!\! \frac{\mu}{2}\!\left \| \mathbf{f} \!\!-\! \mathbf{f}_{t-1} \right \|^2\!\!,
\end{align}
\end{small}
where $\mathbf{s}$, $\mu$ are the Lagrange multiplier and penalty factor, respectively. By introducing $\mathbf{h} = \frac{1}{\gamma}\mathbf{s}$, Eqn. (\ref{equ:STRCFALM}) can be reformulated as
\begin{small}
\setlength\abovedisplayskip{2pt}
\setlength\belowdisplayskip{2pt}
\begin{align}\label{equ:STRCFALM2}
\small
\vspace{-0.1in}
\mathcal{L}(\mathbf{w},\mathbf{g},\mathbf{h}&) = \frac{1}{2}\!\left \| \sum_{d=1}^D \mathbf{x}_t^d \!\ast\!\mathbf{f}^d \!-\!\mathbf{y} \right \|^2 \!+\frac{1}{2} \sum_{d=1}^{D}\left \| \mathbf{w} \!\cdot\! \mathbf{g}^d\right \|^2 \\\nonumber
&+\!\!\frac{\gamma}{2}\! \sum_{d=1}^D\left \| \mathbf{f}^d\!\!-\! \mathbf{g}^d \!\!+\! \mathbf{h}^d\right \|^2 + \frac{\mu}{2}\!\left \| \mathbf{f} - \mathbf{f}_{t-1} \right \|^2\!\!.
\end{align}
\end{small}
The ADMM algorithm is then adopted by alternatingly solving the following subproblems,

\begin{small}
\setlength\abovedisplayskip{-2pt}
\begin{align}\label{equ:ADMM}
\vspace{-0.1in}
\!\!\left\{ \!\!\begin{array}{l}
\!\!{{\bf{f}}^{(i + 1)}} \!\!=\!\! \arg \!\mathop {\min }\limits_{\bf{f}} \!{\left\| {\sum\limits_{d \!=\! 1}^D {{\bf{x}}_t^d} \! *\! {{\bf{f}}^d}\!\! - \!\!{{\bf{y}}}} \right\|^2}\!\!\!\! + \!\gamma {\left\| {{\bf{f}}\! -\! {\bf{g}}\! +\! {\bf{h}}} \right\|^2}\!\! +\!\! \mu {\left\| {{\bf{f}}\!\! -\!\! {{\bf{f}}_{t \!-\! 1}}} \right\|^2}\\
\!\!{{\bf{g}}^{(i + 1)}} = \arg \mathop {\min }\limits_{\bf{g}} \sum\limits_{d = 1}^D {{{\left\| {{\bf{w}} \cdot {{\bf{g}}^d}} \right\|}^2} + \gamma {{\left\| {{\bf{f}} - {\bf{g}} + {\bf{h}}} \right\|}^2}} \\
\!\!{{\bf{h}}^{\left( {i + 1} \right)}} = {{\bf{h}}^{\left( i \right)}} + {{\bf{f}}^{\left( {i + 1} \right)}} - {{\bf{g}}^{\left( {i + 1} \right)}}.
\end{array} \right.
\end{align}
\end{small}

We detail the solution to each subproblem as follows:
\noindent
\textbf{Subproblem $\mathbf{f}$}: Using the Parseval's theorem, the first row of Eqn. (\ref{equ:ADMM}) can be rewritten in the Fourier domain as
\begin{small}
\begin{align}\label{equ:SubProblemf}
\!\! \arg \!\mathop {\min }\limits_{\hat{\bf{f}}} \!{\left\| {\sum\limits_{d \!=\! 1}^D {{\hat{\bf{x}}}_t^d} \! \cdot\! {\hat{{\bf{f}}}^d}\!\! - \!\!{\hat{{\bf{y}}}}} \right\|^2}\!\!\!\! + \!\gamma {\left\| {{\hat{\bf{f}}}\! -\! \hat{{\bf{g}}}\! +\! {\hat{\bf{h}}}} \right\|^2}\!\! +\!\! \mu {\left\| {{\hat{\bf{f}}}\!\! -\!\! {{\hat{\bf{f}}}_{t \!-\! 1}}} \right\|^2},
\end{align}
\end{small}
where $\hat{\mathbf{f}}$ denotes the discrete Fourier transform (DFT) of the filter $\mathbf{f}$.
From Eqn. (\ref{equ:SubProblemf}), we can see that the $j$-th element of the label $\hat{\mathbf{y}}$ only depends on the $j$-th element of the filter $\hat{\mathbf{f}}$ and sample $\hat{\mathbf{x}}_t$ across all $D$ channels.
Denote by $\mathcal{V}_j(\mathbf{f}) \in \mathbb{R}^D$ the vector consisting of the $j$-th elements of $\mathbf{f}$ along all $D$ channels.
Eqn. (\ref{equ:SubProblemf}) can be further decomposed into $MN$ subproblems, where each of them is defined as
\begin{small}
\begin{align}\label{equ:SubProblemf2}
\arg\!\!\min\limits_{\mathcal{V}_j(\hat{\mathbf{f}})}&\left \| \mathcal{V}_j(\hat{\mathbf{x}}_t)^\top \mathcal{V}_j(\hat{\mathbf{f}}) \!\!-\!\! {\hat{y}_j} \right \|^2 \!+ \!\!\mu\!\left \| \mathcal{V}_j(\hat{\mathbf{f}}) \!\!-\!\! \mathcal{V}_j(\hat{\mathbf{f}}_{t-1}) \right \|^2\\\nonumber
&+\!\gamma\! \left \| \mathcal{V}_j(\hat{\mathbf{f}})\! -\! \mathcal{V}_j(\hat{\mathbf{g}})\! +\! \mathcal{V}_j(\hat{\mathbf{h}}) \right \|^2\!\!\! .
\end{align}
\end{small}
Taking the derivative of Eqn. (\ref{equ:SubProblemf2}) be zero, we can get the closed-form solution for $\mathcal{V}_j(\hat{\mathbf{f}})$,
\begin{small}
\begin{align}\label{equ:SubProblemf3}
\mathcal{V}_j(\hat{\mathbf{f}}) = (\mathcal{V}_j(\hat{\mathbf{x}}_t)\mathcal{V}_j(\hat{\mathbf{x}}_t)^\top+(\mu+\gamma)I)^{-1} \mathbf{q},
\end{align}
\end{small}
where the vector $\mathbf{q}$ takes the form as
$\mathbf{q} = \mathcal{V}_j(\hat{\mathbf{x}}_t){\hat{y}_j} + \gamma \mathcal{V}_j(\hat{\mathbf{g}})- \gamma\mathcal{V}_j(\hat{\mathbf{h}}) + \mu \mathcal{V}_j(\hat{\mathbf{f}}_{t-1})$.
Since $\mathcal{V}_j(\hat{\mathbf{x}}_t)\mathcal{V}_j(\hat{\mathbf{x}}_t)^\top$ is rank-1 matrix, Eqn. (\ref{equ:SubProblemf3}) can be solved with the Sherman-Morrsion formula \cite{Pedersen2008Cookbook}, and we have
\begin{small}
\begin{align}\label{equ:SubProblemf4}
\mathcal{V}_j(\hat{\mathbf{f}}) = \frac{1}{\mu+\gamma}(I- \frac{\mathcal{V}_j(\hat{\mathbf{x}}_t)\mathcal{V}_j(\hat{\mathbf{x}}_t)^\top}{\mu+\gamma+\mathcal{V}_j(\hat{\mathbf{x}}_t)^\top\mathcal{V}_j(\hat{\mathbf{x}}_t)})\mathbf{q}.
\end{align}
\end{small}
Note that Eqn. (\ref{equ:SubProblemf4}) only contains vector multiply-add operation and thus can be computed efficiently. $\mathbf{f}$ can be further obtained by the inverse DFT of $\hat{\mathbf{f}}$.\\
\noindent
\textbf{Subproblem $\mathbf{g}$}: From the second sub-equation of Eqn. (\ref{equ:ADMM}), each element of $\mathbf{g}$ can be computed independently,
and thus the closed-form solution of $\mathbf{g}$ can be computed by,
\begin{small}
\begin{align}\label{equ:SubProblemG}
\mathbf{g} = (\mathbf{W}^\top\mathbf{W}+\gamma I)^{-1}(\gamma\mathbf{f}+\gamma\mathbf{h}).
\end{align}
\end{small}
where $\mathbf{W}$ represents the $DMN \times DMN$ diagonal matrix {concatenated with $D$ diagonal matrices $\text{Diag}(\mathbf{w})$.}\\
\textbf{Updating stepsize parameter $\gamma$}: The stepsize parameter $\gamma$ is updated as in Eqn. (\ref{equ:penaltyFactor}),
\begin{align}\label{equ:penaltyFactor}
\gamma^{(i+1)} = min(\gamma^{\max}, \rho\gamma^{(i)}),
\end{align}
where $\gamma^{\max}$ denotes the maximum value of $\gamma$ and the scale factor $\rho$.\\
%
%
{\textbf{Complexity Analysis.}
Since Eqn. (\ref{equ:SubProblemf}) is separable in each pixel location, we should solve $MN$ subproblems and each is a system of linear equations with $D$ variables. With Sherman-Morrison formula, each system can be solved in $\mathcal{O}(D).$ \footnote{please refer to \cite{zuo2016learning} for more details.}Thus, the complexity of solving $\hat{\mathbf{f}}$ is
$\mathcal{O}(DMN)$. Taking the DFT and inverse DFT into account, the complexity of solving $\mathbf{f}$ is
$\mathcal{O}(DMN\text{log}(MN))$. And the computational cost for $\mathbf{g}$ is $\mathcal{O}(DMN)$.}
Hence, the overall cost of our algorithm is $\mathcal{O}(DMN\log(MN)N_{I})$, where $N_I$ represents the maximum number of iterations.
In addition, compared with SRDCF, our ADMM algorithm does not need a start-up time to initialize the CFs in the first frame.\\
\textbf{Convergence.} Note that the STRCF model is convex, and each sub-problem in ADMM algorithm has closed-form solution. Therefore, it satisfies the Eckstein-Bertsekas condition~\cite{Eckstein1992On}, and is guaranteed to converge to global optimum. In addition, We empirically find that the proposed ADMM can converge within {2} iterations on most of the sequences, and thus $N_I$ is set to {2} for efficiency.

\section{Experimental Results}
In this section, we first compare our STRCF with the state-of-the-art trackers in terms of both hand-crafted and CNN features on the OTB-2015 dataset.
Then, we analyze the impacts of the temporal regularization and hyper-parameter $\mu$ on tracking performance using OTB-2015.
Finally, we conduct comparative experiments on Temple-Color and  VOT-2016 benchmarks.

Following the settings in SRDCF \cite{danelljan2015learning}, we crop the square region centered at the target,
in which the side length of the region is $\sqrt{5WH}$ ($W$ and $H$ represent the width and height of the target, respectively).
Then we extract {HOG, CN \cite{Danelljan2014Adaptive} and CNN features} for the image region.
The features are further weighted by a cosine window to reduce the boundary discontinuities.
{As for the ADMM algorithm, we set the hyper-parameter in Eqn. (\ref{equ:STRCF}) to $\mu = 16$ throughout all the experiments.
The initial stepsize parameter $\gamma^{(0)}$, the maximum value $\gamma^{\max}$ and scale factor $\rho$ are set to $10$, $100$ and $1.2$, respectively.}
Our STRCF is implemented with Matlab 2017a and all the experiments are run on a PC equipped with Intel i7 7700 CPU, 32GB RAM {and a single NVIDIA GTX 1070 GPU.}
The source code of our tracker is publicly available at \url{https://github.com/lifeng9472/STRCF}.

\subsection{The OTB-2015 benchmark}

The OTB-2015 benchmark \cite{wu2015object} is a popular tracking dataset which consists of 100 fully annotated video sequences with 11 different attributes,
such as abrupt motion, illumination variation, scale variation and motion blurring.
We evaluate the trackers based on the One Pass Evaluation (OPE) protocol provided in \cite{wu2015object},
where overlap precision (OP) metric is employed by calculating the bounding box overlaps exceeding 0.5 in a sequence.
Besides, we also provide the overlap success plots containing the OP metric over a range of thresholds.

\begin{table*}[!htb]
\centering
\setlength{\belowcaptionskip}{0.2cm}
\scalebox{0.55}{
\begin{tabular}{ccccccccccccccc}
\toprule
 & SRDCF \cite{danelljan2015learning} & BACF \cite{kiani2017learning} & ECO-HC \cite{Danelljan2016ECO}& SRDCFDecon \cite{Danelljan2016Adaptive} &Staple \cite{bertinetto2015staple}&Staple+CA\cite{cfcatracking} & SAMF+AT \cite{Bibi2016Target} & SAMF \cite{li2014scale} & MEEM \cite{zhang2014meem}  & DSST \cite{danelljan2016discriminative}  & KCF \cite{henriques2015high} & STRCF (HOG) & STRCF (HOGCN)\\
\midrule
Mean OP &72.7& 77.5&{\color{red}79.6} &{\color{green}77}& 71& 73.8 & 68  & 64.4 & 62.3 & 62.2 & 55.5 & {\color{blue}79.2} & {\color{red}79.6} \\
FPS & 5.8 & 26.7 & {\color{green}42}&2.0 & {\color{blue}76.6}& 35.3 & 2.2 & 23.2 & 22.4 & 20.4 &  {\color{red}171.8}& 31.5& 24.3   \\
\bottomrule
\end{tabular}}
\vspace{-0.08in}
\caption{\small{The mean OP (in \%) and FPS results of trackers with hand-crafted features on OTB-2015. The best three results are shown in {\color{red}red}, {\color{blue}blue} and {\color{green}green} fonts, respectively.}}
\label{tab:HandMeanOP}
\vspace{-0.08in}
\end{table*}

\begin{table*}[!htb]
\centering
\setlength{\abovecaptionskip}{0.4cm}
\setlength{\belowcaptionskip}{0.2cm}
\scalebox{0.7}{
\begin{tabular}{ccccccccccccccc}
\toprule
         & ECO \cite{Danelljan2016ECO} & DeepSRDCF \cite{danelljan2015learning} & SiameseFC \cite{bertinetto2016fully}& FCNT \cite{wang2015visual}& HDT \cite{qi2016hedged}& MSDAT \cite{wang2017robust} & HCF \cite{ma2015hierarchical}& C-COT \cite{Danelljan2016CCOT}& CF-Net \cite{valmadre2017end}& DeepSTRCF \\
\midrule
Mean OP & {\color{red}85.5} & 76.8 & 71 & 67.1 & 65.8 & 65.6 & 65.6 & {\color{green}82.7} & 73 & {\color{blue}84.2}\\
FPS & 9.8 & 0.2 & {\color{red}83.7}   & 1.2  & 2.7   & {\color{green}25} & 10.2 & 0.8 & {\color{blue}78.4}& 5.3 \\
\bottomrule
\end{tabular}}
\vspace{-0.08in}
\caption{\small{The OP metric (in \%) and FPS results of trackers with deep features on OTB-2015. The best three results are shown in {\color{red}red}, {\color{blue}blue} and {\color{green}green} fonts, respectively.}}
\label{tab:DeepMeanOP}
\vspace{-0.12in}
\end{table*}

We compare STRCF with 20 state-of-the-art trackers, including
trackers using hand-crafted features (\ie SRDCF \cite{danelljan2015learning}, BACF \cite{kiani2017learning}, ECO-HC \cite{Danelljan2016ECO}, SRDCFDecon \cite{Danelljan2016Adaptive}, Staple \cite{bertinetto2015staple}, Staple+CA\cite{cfcatracking}, SAMF+AT \cite{Bibi2016Target}, DSST \cite{danelljan2016discriminative}, SAMF \cite{li2014scale}, MEEM \cite{zhang2014meem} and KCF \cite{henriques2015high}) and using CNN features (\ie ECO \cite{Danelljan2016ECO}, DeepSRDCF \cite{danelljan2015convolutional}, HCF \cite{ma2015hierarchical}, HDT \cite{qi2016hedged}, C-COT \cite{Danelljan2016CCOT}, FCNT \cite{wang2015visual}, SiameseFC \cite{bertinetto2016fully}, CF-Net \cite{bertinetto2016fully} and MSDAT \cite{wang2017robust}).
Note that we employ the publicly available codes or results provided by the authors for fair comparison.

\begin{spacing}{-0.5}
\end{spacing}

\subsubsection{Comparison with hand-crafted based trackers}

We compare the proposed STRCF with other state-of-the-art trackers using hand-crafted features.
Table \ref{tab:HandMeanOP} gives the results of the mean OP and FPS on OTB-2015.
As shown in Table \ref{tab:HandMeanOP}, STRCF performs significantly better than most of the competing trackers except ECO-HC and surpasses its counterpart SRDCF by 6.9\%.
We owe these significant improvements to the introduction of the temporal regularization.
STRCF is also superior to the SRDCFDecon tracker which follows the SRDCF work and addresses the corrupted sample problem by re-weighting the samples in the training set.
It indicates that the introduction of temporal regularization is more helpful than multiple samples training with explicit sample re-weighting.
{Besides, our method also outperforms the recent CF-based trackers: BACF~\cite{Galoogahi2017Learning}, SAMF+AT~\cite{Bibi2016Target} and Staple+CA~\cite{cfcatracking}.}
Overall, the only tracker performing comparably with STRCF on OTB-2015 is the ECO-HC \cite{Danelljan2016ECO}.
{It is worth noting that ECO-HC adopts the Gaussian Mixture Model (GMM)-based generative sample space method to reduce the number of samples for training, and employs continuous convolution and factorized convolution for boosting the performance.
In contrast, even our STRCF does not consider continuous convolution and factorized convolution techniques,
it still yields favorable performance against the competing trackers.}

In addition, we also report the tracking speed (FPS) comparison on OTB-2015 dataset in Table \ref{tab:HandMeanOP}.
One can see that STRCF (HOGCN) runs at 24.3 FPS and is nearly 4.2$\times$ than its counterpart SRDCF (5.8 FPS),
validating the high efficiency of the proposed ADMM over the SRDCF solver (\ie the Gauss-Seidel algorithm).
STRCF (HOG) using HOG feature performs even faster and obtains a real-time speed of 31.5 FPS,
which is 1.2$\times$ faster than recent BACF tracker.

\begin{figure}[!htbp]
\centering
\setlength{\belowcaptionskip}{-0.4cm}
\subfloat[]{\label{fig:OPfig(1)}
  \includegraphics[width=0.23\textwidth]{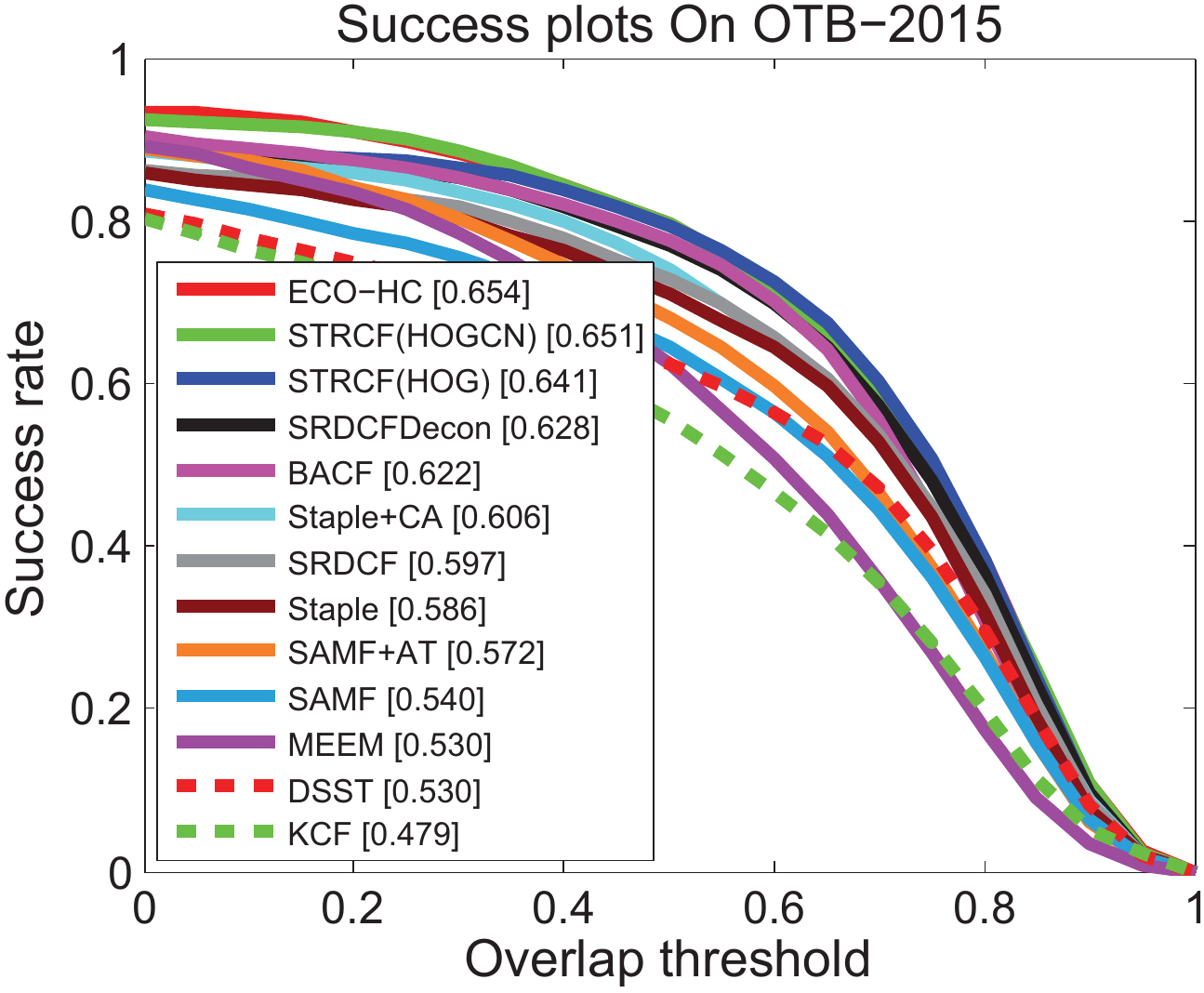}}
\subfloat[]{\label{fig:OPfig(2)}
  \includegraphics[width=0.23\textwidth]{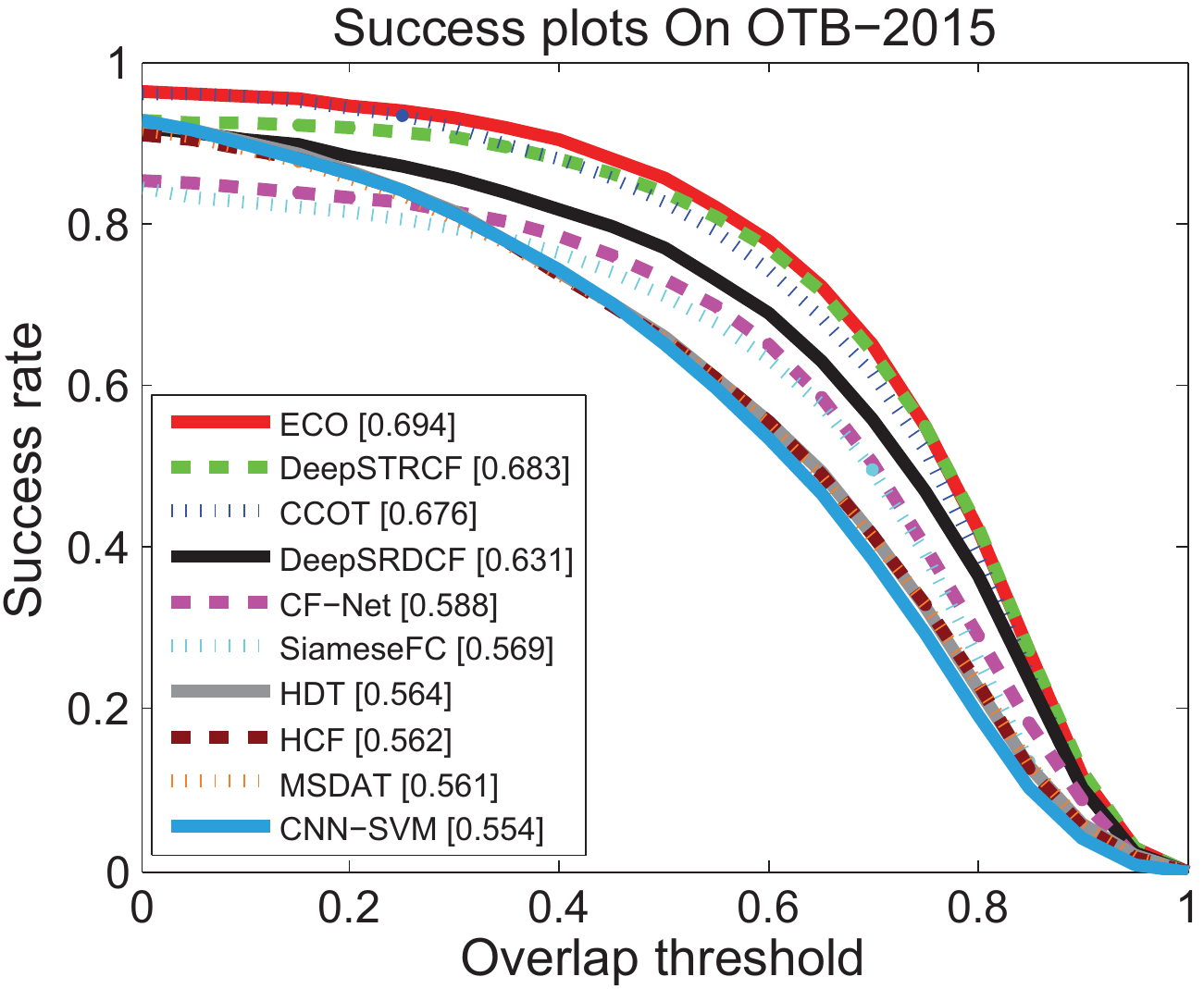}}
\caption{A comparison of the overlap success plots with the state-of-the-art trackers on OTB-2015 dataset. (a) Trackers with hand-crafted features. (b) Trackers with deep features.}
\label{fig:OPfig}
\end{figure}

Next, we provide the overlap success curves of the competing trackers with the hand-crafted features on OTB-2015 dataset, which is ranked using the Area-Under-the-Curve (AUC) score.
As shown in Fig. \ref{fig:OPfig(1)}, our STRCF achieves an AUC score of 65.1\% and ranks the second best performance among all the trackers.
Similar to the mean OP results, STRCF also outperforms its counterparts SRDCF and SRDCFDecon by a gain of 5.4\% and 2.3\%, respectively.

Finally, we perform qualitative evaluation of different trackers on several video sequences.
For clearer visualization, we show the results of STRCF and 4 state-of-the-art trackers based on hand-crafted features, including ECO-HC~\cite{Danelljan2016ECO}, BACF~\cite{kiani2017learning}, SRDCF~\cite{danelljan2015learning} and SRDCFDecon~\cite{Danelljan2016Adaptive}.
The tracking results on on 6 video sequences are shown in Fig. \ref{fig:Visual}.
One can note that, with the introduction of temporal regularization, the proposed STRCF performs favorably against the state-of-the-art hand-crafted trackers.
\begin{figure*}
\centering
\includegraphics[width=1\textwidth]{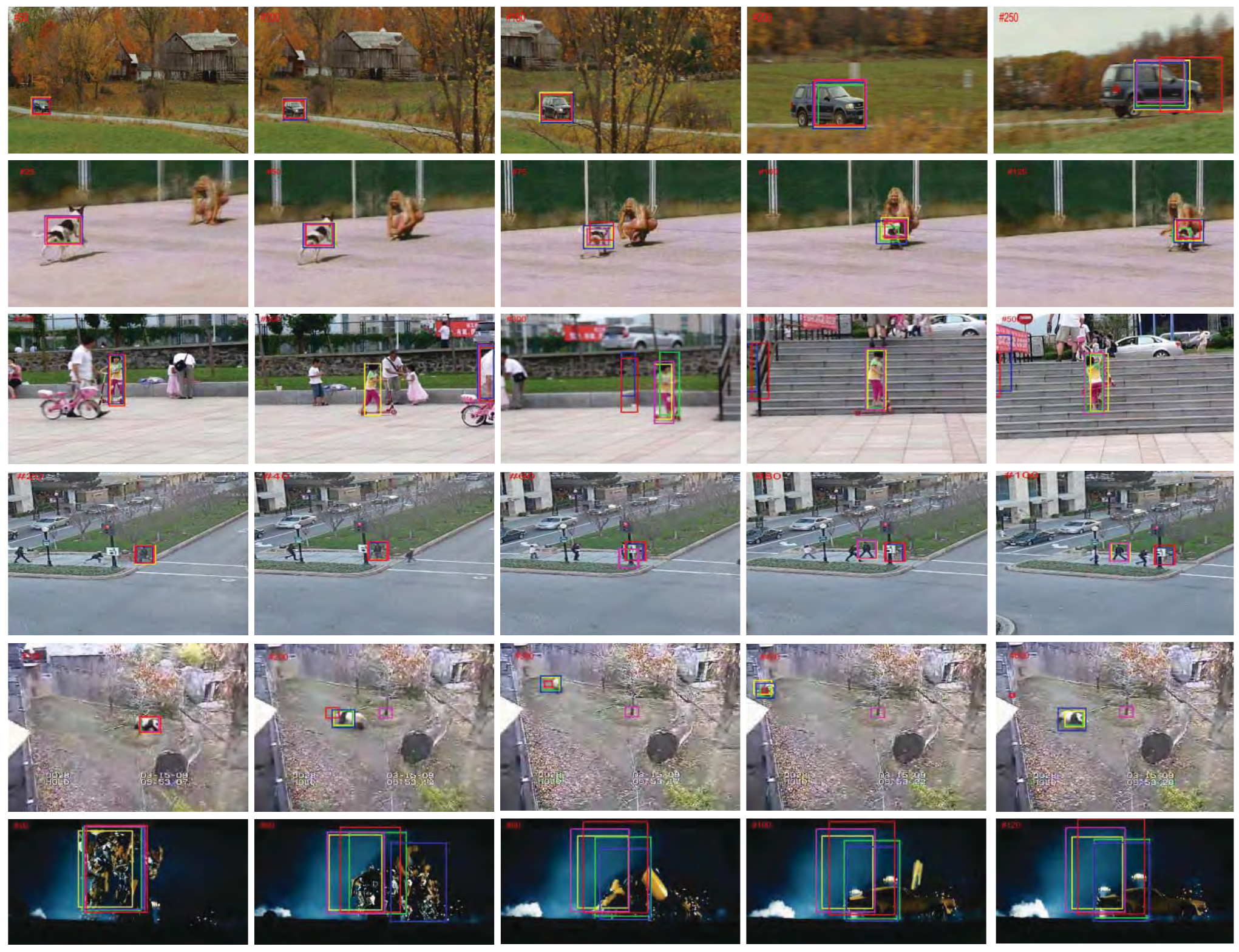}
\caption{Qualitative evaluation on 6 video sequences (\ie \emph{CarScale}, \emph{Dog}, \emph{Girl2},  \emph{Human3}, \emph{Panda} and \emph{Trans}). We show the results of {\color{green}STRCF}, {\color{yellow}ECO-HC}, {\color{blue}BACF}, {\color{red}SRDCF} and {\color{cyan}SRDCFDecon} with different colors, respectively.}
\label{fig:Visual}
\end{figure*}

\subsubsection{Video Attribute Based Comparison}

In this section, we perform quantitative analysis of the total 11 video attributes on the OTB-2015 dataset.
Our STRCF outperforms most of the competing trackers except ECO-HC on all the attributes.
Due to the page limits, here we only provide the overlap success plots of 4 attributes in Fig. \ref{fig:Attributes} and the remaining results can be found in the supplementary material.

In the case of out of view (OV) and occlusion (OCC), the target always encounters with partial or fully disappearance from the camera, which leads to an adverse impact on model updating.
Trackers using multiple samples training with naive sample weighting strategy (\ie SRDCF) or linear interpolation updating (\ie Staple and SAMF+AT) suffer from significant degradation because of over-fitting to the recent samples.
Benefited from the temporal regularization, our STRCF can adaptively make the balance between updating the CFs with the latest samples and
keeping close to the previously learned CFs, and thus is robust to such kinds of variations.
In particular, STRCF achieves remarkable improvements over its baseline SRDCF, i.e., 14.5\% and 5.7\% gains on these two attributes, respectively.
And it also outperforms the SRDCFDecon tracker by an AUC score of 9.3\% and 2.1\% on these two attributes.
As for the In-plane/Out-of-plane rotation attributes, STRCF also performs better than most of the trackers and is superior to the baseline SRDCF by 5.8\% and 7.6\%, respectively.
\begin{figure*}[htbp]
\setlength{\abovecaptionskip}{0.2cm}
\setlength{\belowcaptionskip}{-0.4cm}
\centering
\subfloat{%
  \includegraphics[width=0.24\textwidth]{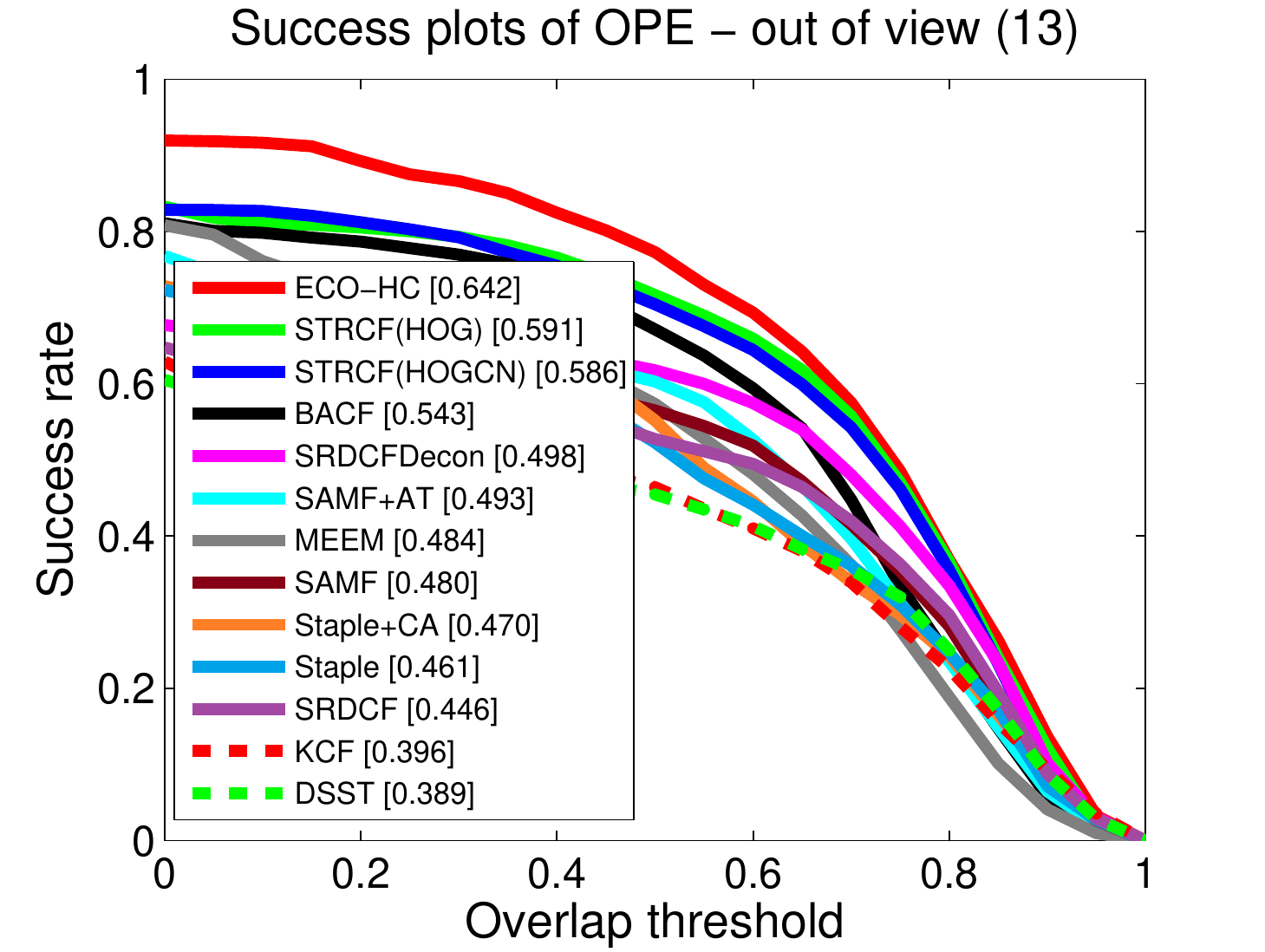}}\
\subfloat{%
  \includegraphics[width=0.24\textwidth]{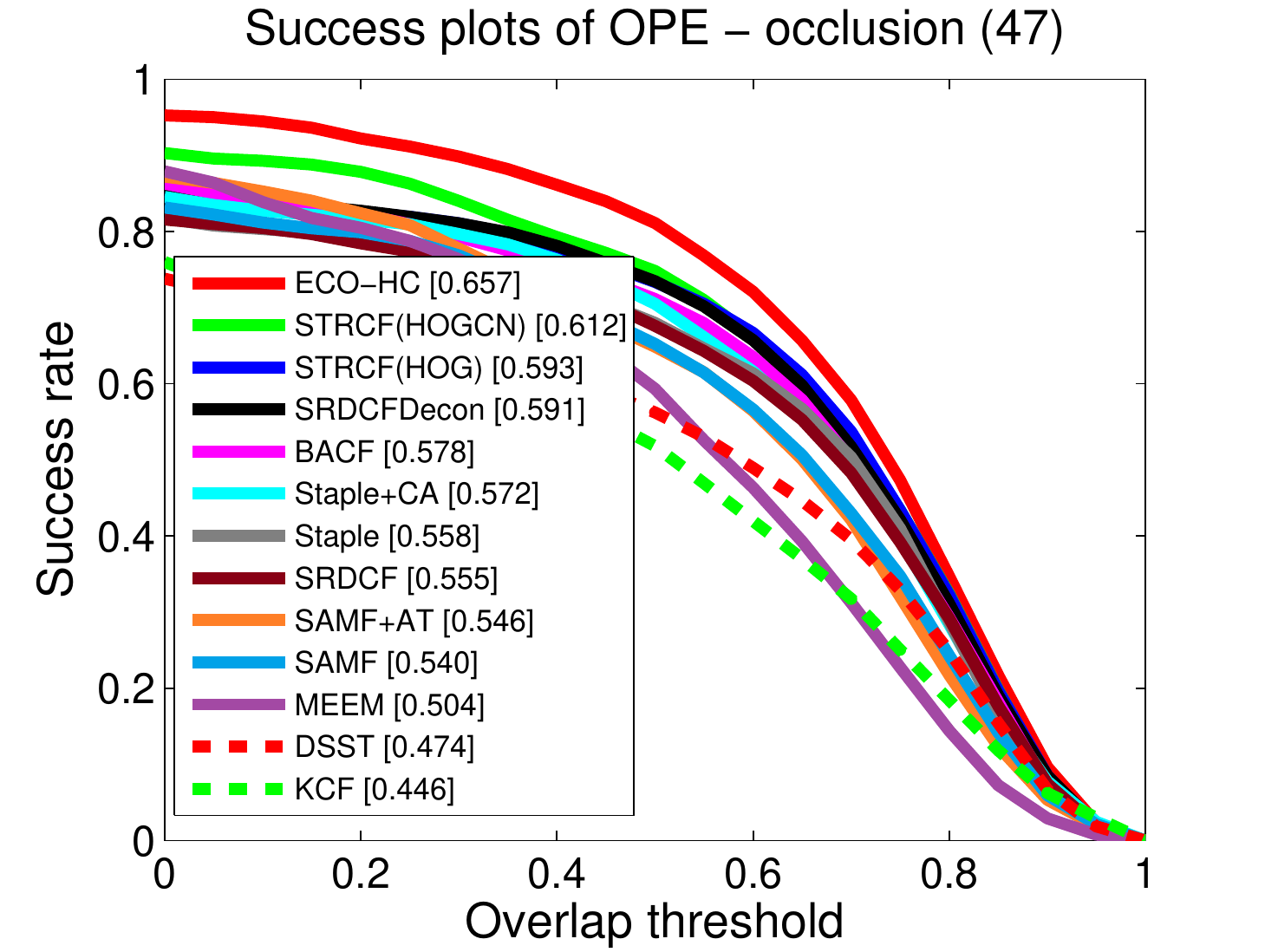}}\
\subfloat{%
  \includegraphics[width=0.24\textwidth]{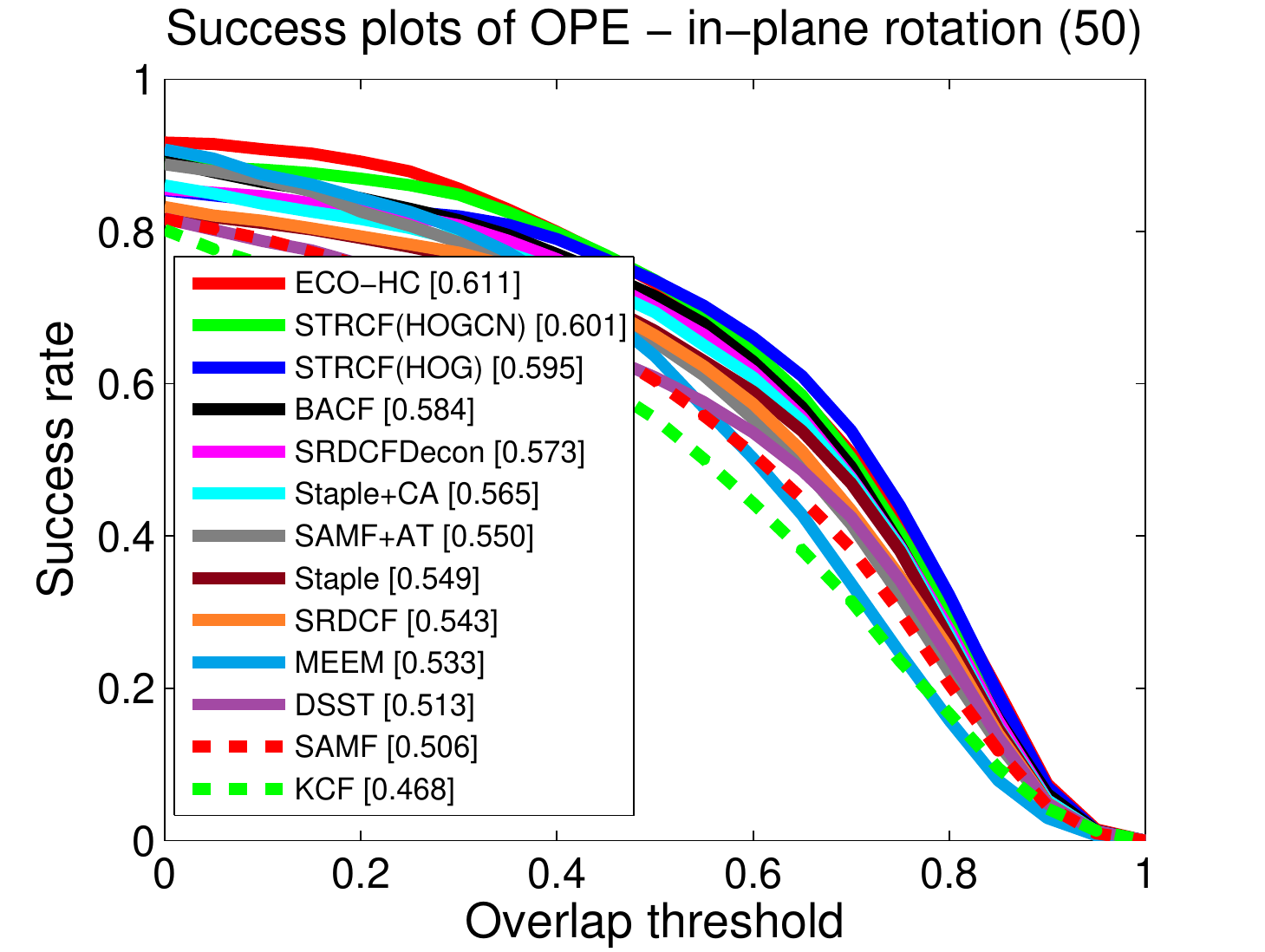}}\
\subfloat{%
  \includegraphics[width=0.24\textwidth]{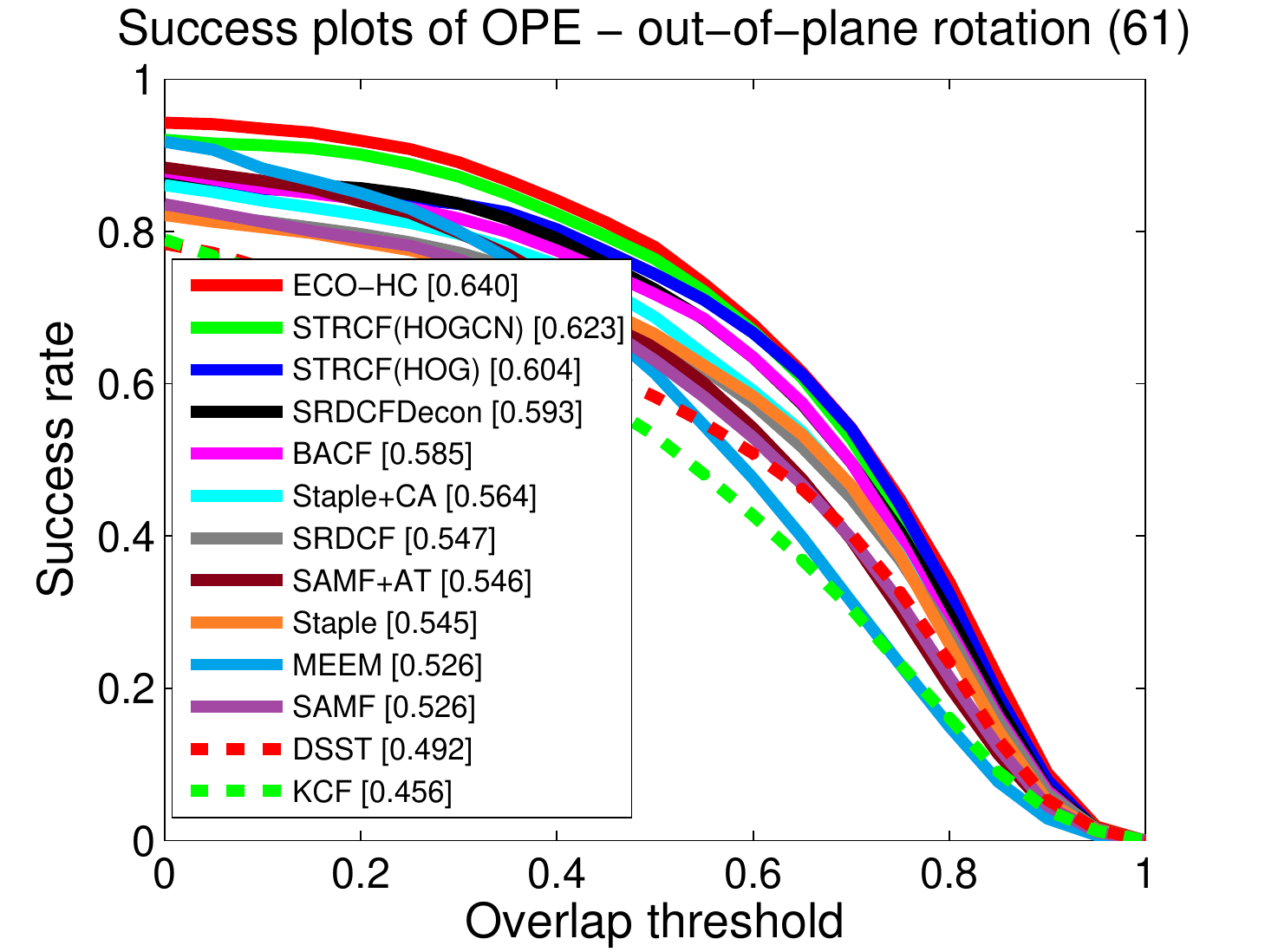}}\
\caption{\small{The overlap success plots of the competing trackers with 4 video attributes on the OTB-2015 dataset.}}
\label{fig:Attributes}
\end{figure*}
\begin{figure}[htbp]
\setlength{\abovecaptionskip}{0.2cm}
\setlength{\belowcaptionskip}{-0.2cm}
\centering
\subfloat[]{\label{fig:Ablative}
  \includegraphics[width=0.225\textwidth]{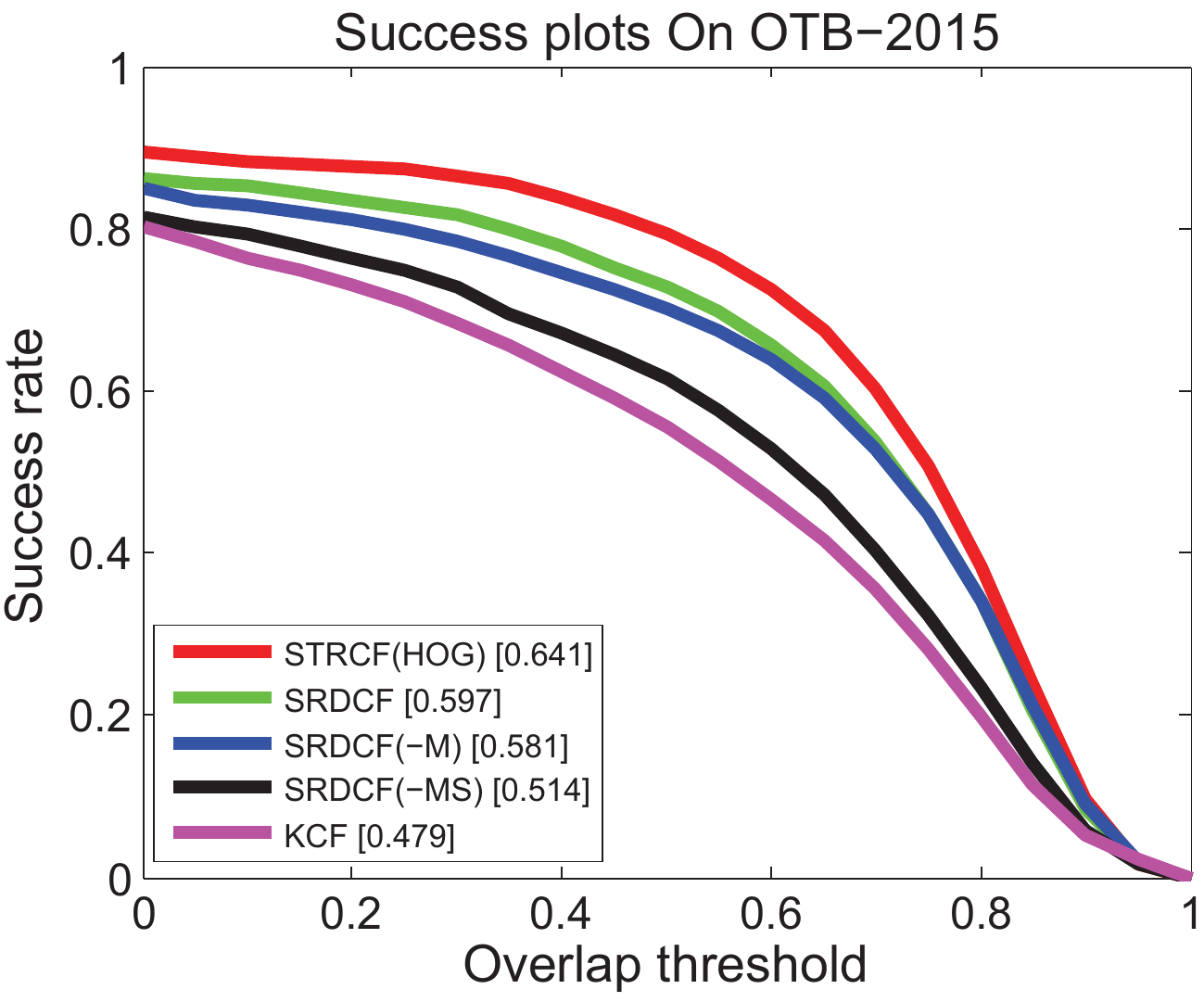}}\
\subfloat[]{\label{fig:CFvars}
\includegraphics[width=0.23\textwidth]{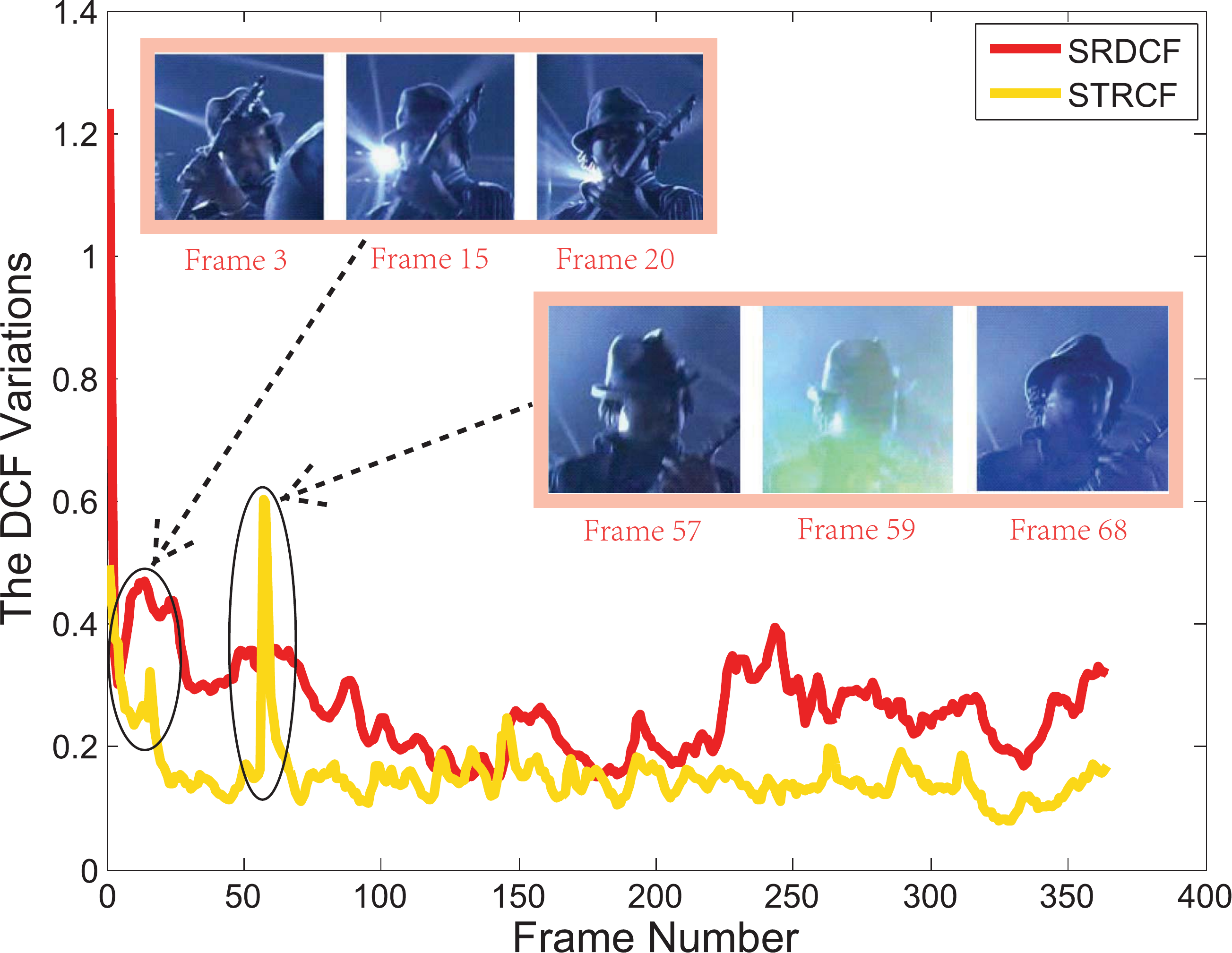}}\
\caption{\small{Ablative study on the STRCF method. (a) The overlap success plot of the SRDCF variants  and our STRCF on OTB-2015. (b) The visualization results of temporal CF variation against frames on sequence \emph{Shaking.}}}
\end{figure}
{\subsubsection{Comparison with deep feature-based trackers}

To further assess STRCF, we follow the settings in C-COT \cite{Danelljan2016CCOT}, and combine the outputs of \emph{conv}3 layer from VGG-M network \cite{simonyan2014very} with HOGCN features for STRCF training (we name it as DeepSTRCF for simplicity).
Using mean OP and speed as performance metrics, Table \ref{tab:DeepMeanOP} compares DeepSTRCF with the state-of-the-art trackers based on deep features on OTB-2015.
One can see that DeepSTRCF achieves a mean OP of 84.2\% and
performs much better than the SRDCF with CNN features (\ie DeepSRDCF) by a gain of 7.4\%, demonstrating the effectiveness of the temporal regularization.
It even outperforms than C-COT with both spatial regularization and continuous convolution by a gain of 1.2\% on OTB-2015.
In terms of the tracking speed, the best performance belongs to SiameseFC (83.7 FPS), followed by CF-Net (78.4 FPS) and MSDAT (25 FPS), while DeepSTRCF runs at 5.3 FPS.
The higher speed of these trackers, however, comes at the cost of much lower accuracy in comparison to STRCF.
Furthermore, We also provide the overlap success curves of the competing trackers in Fig. \ref{fig:OPfig(2)}.
One can see that DeepSTRCF ranks the second and outperforms DeepSRDCF with a margin of 5.2\% on OTB-2015.}
\begin{table*}[htbp]
\setlength{\abovecaptionskip}{0.1cm}
\setlength{\belowcaptionskip}{-0.2cm}
\centering
\scalebox{0.58}{
\begin{tabular}{ccccccccccccccccc}
\toprule
 & DSST \cite{danelljan2016discriminative} &ECO \cite{Danelljan2016ECO}& Staple \cite{bertinetto2015staple}& MDNet\_N \cite{nam2016mdnet} & TCNN \cite{nam2016modeling}& SRDCF \cite{danelljan2015learning}& BACF \cite{kiani2017learning}& SRDCFDecon \cite{Danelljan2016Adaptive}& DeepSRDCF \cite{danelljan2015convolutional}& ECO-HC \cite{Danelljan2016ECO} & STRCF &DeepSTRCF\\
\midrule
EAO &0.181&0.375&0.295&0.257&0.325&0.247&0.223&0.262&0.276&0.322&0.279&0.313\\
\midrule
Accuracy &0.5&0.53&0.54&0.53&0.54&0.52&0.56&0.53&0.51&0.54&0.53&0.55\\
\midrule
Robustness & 2.72&0.73&1.35&1.2&0.96&1.5&1.88&1.42&1.17&1.08&1.32&0.92\\
\bottomrule
\end{tabular}}
\caption{\small{A comparison with the state-of-the-art trackers on VOT-2016 dataset.}}
\label{tab:VOT2016}
\end{table*}

\subsection{Internal Analysis of the proposed approach}

\subsubsection{Impacts of the Temporal regularization}

In this section, we investigate the impacts of the temporal regularization on the proposed STRCF approach using the OTB-2015 dataset.
Fig. \ref{fig:Ablative} gives the overlap success plot of different SRDCF variants (discussed in Section~\ref{sec:Intro}) and our STRCF.
Compared with the KCF method, we can see that the introduction of scale estimation (\ie SRDCF(-MS)) and spatial regularization (\ie SRDCF(-M)) can boost the performance by 3.5\% and 6.7\%, respectively.
Besides, SRDCF also outperforms SRDCF(-M) by 1.6\% with the coupling of DCF learning and model updating.
However, when incorporating the temporal regularization into SRDCF(-M) formulation, our STRCF can bring notable improvements over both SRDCF(-M) and SRDCF with a gain of 6\% and 4.4\%, respectively.
This can be explained by the merits of online PA on adaptively balancing the tradeoff between aggressive and passive model updating.

To further illustrate the differences of STRCF and SRDCF on model learning, we visualize the temporal CF variation (\ie $\frac{\left \| \mathbf{f}_t\!- \!\mathbf{f}_{t-1} \right \|^{2}}{z}$, where $z$ is the normalization factor) against frames on sequence \emph{Shaking} in Fig. \ref{fig:CFvars}.
From it we can draw the following conclusions:
(1) Compared with the SRDCF tracker, our STRCF \emph{passively} updates the CFs in most frames with small appearance variations, thus leading to more robust DCF variations.
(2) While SRDCF suffers from slow appearance variations (\ie occlusion in the $3\!\!\sim\!\!20$-th frames), our STRCF is dominated by the \emph{passive} model learning and thus insensitive to these variations.
(3) In the case of sudden appearance variations (\ie the illumination changes in the $58\!\!\sim\!\!68$-th frames), STRCF can benefit from the \emph{aggressive} model learning and better adapt to these situations than SRDCF.
{It should be noted that these phenomena are ubiquitous in various video attributes, and the visualizations of the temporal CF variations on more videos are given in Fig. \ref{fig:ALLCFVars}.}
In summary, with the introduction of the temporal regularization, our STRCF can provide a more robust appearance model than SRDCF, thereby leading to superior performance.
\begin{figure*}
\centering
\subfloat[\emph{Bolt}]{\label{fig:Bolt}
  \includegraphics[width=0.46\textwidth,angle=0]{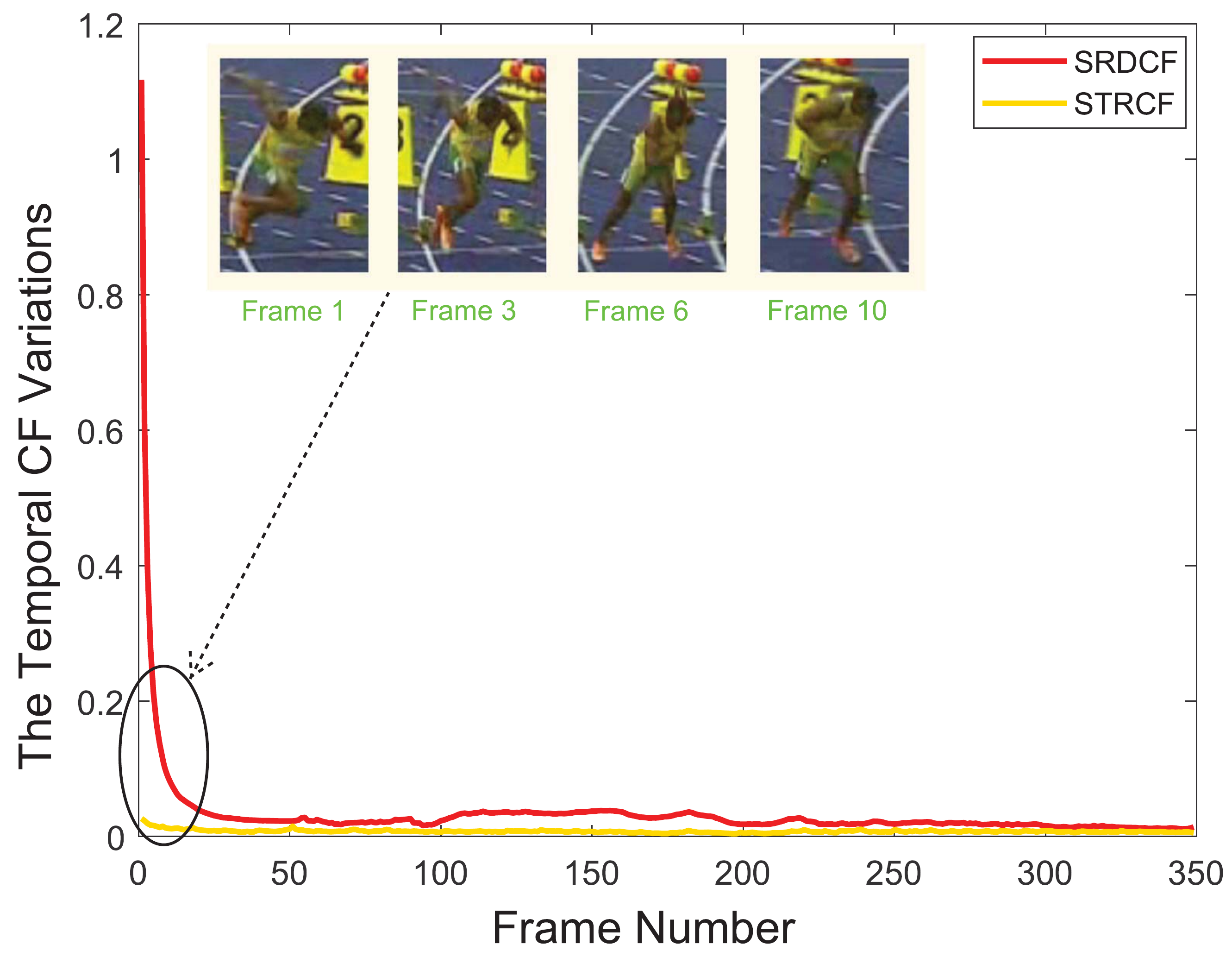}}\
\subfloat[\emph{Bolt2}]{\label{fig:Bolt2}
  \includegraphics[width=0.45\textwidth,angle=0]{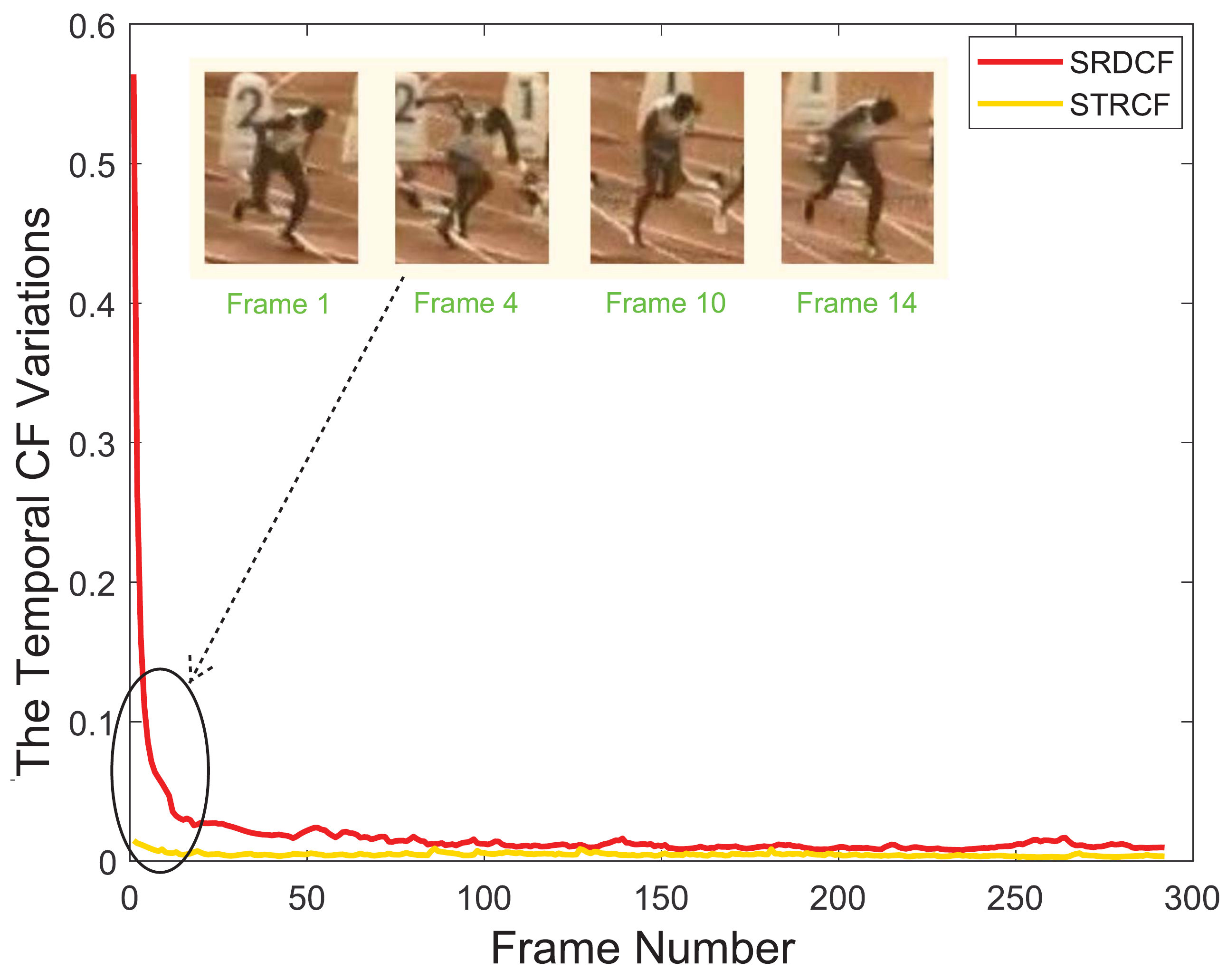}}\\[1mm]
\subfloat[\emph{Panda}]{\label{fig:Panda}
  \includegraphics[width=0.46\textwidth,angle=0]{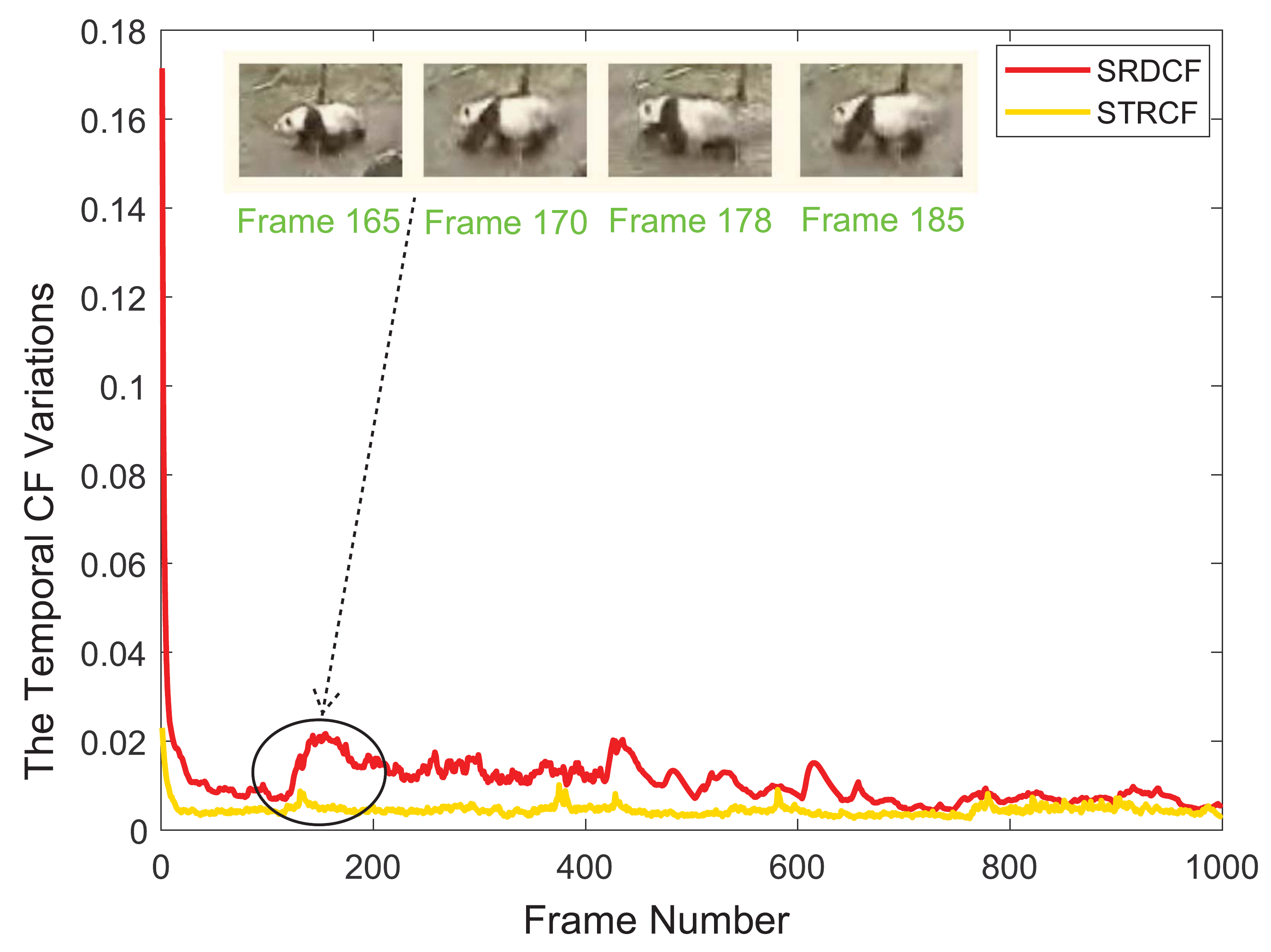}}\
\subfloat[\emph{DragonBaby}]{\label{fig:DragonBaby}
  \includegraphics[width=0.44\textwidth,angle=0]{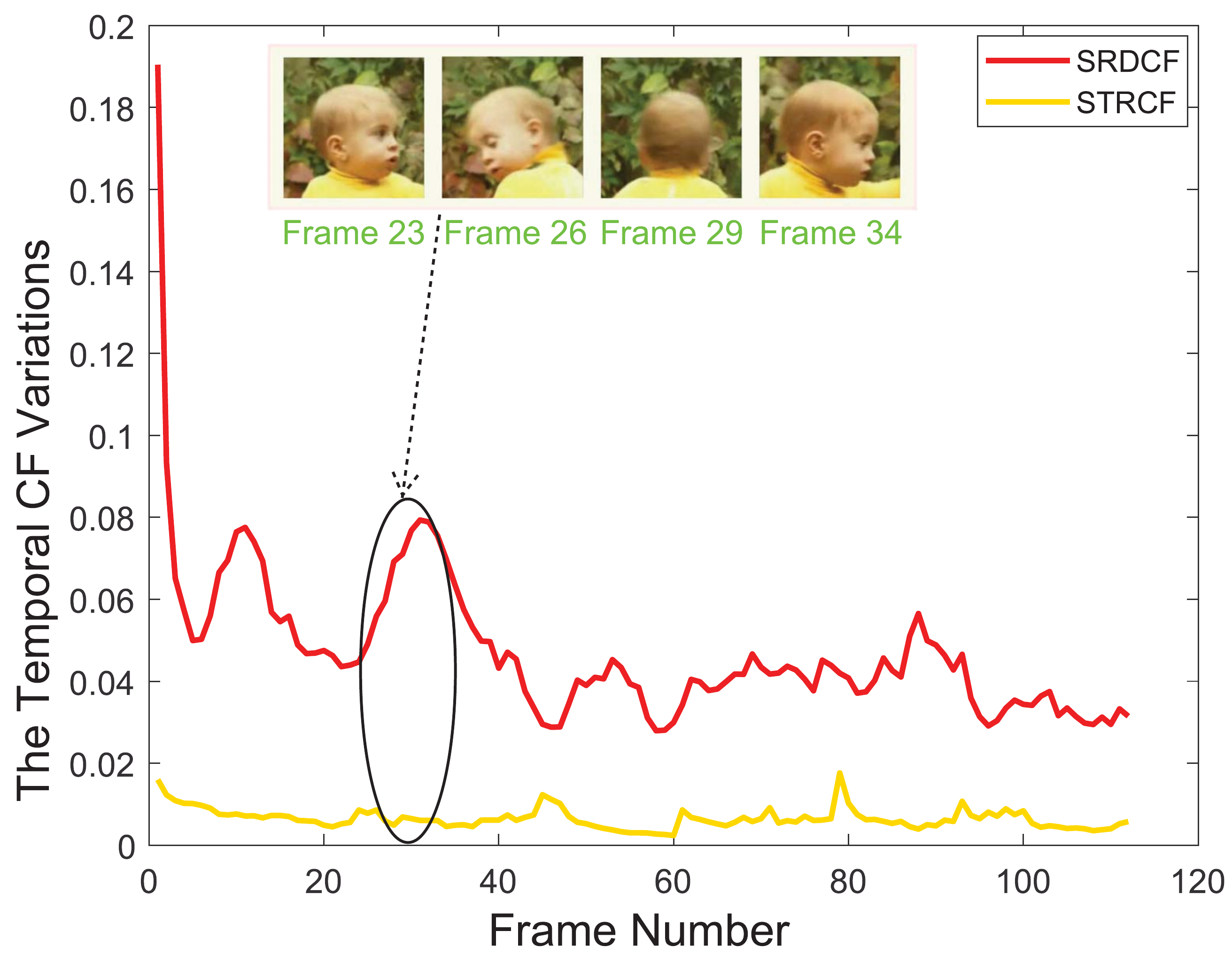}}\\[1mm]
\subfloat[\emph{Football1}]{\label{fig:Football1}
  \includegraphics[width=0.46\textwidth,angle=0]{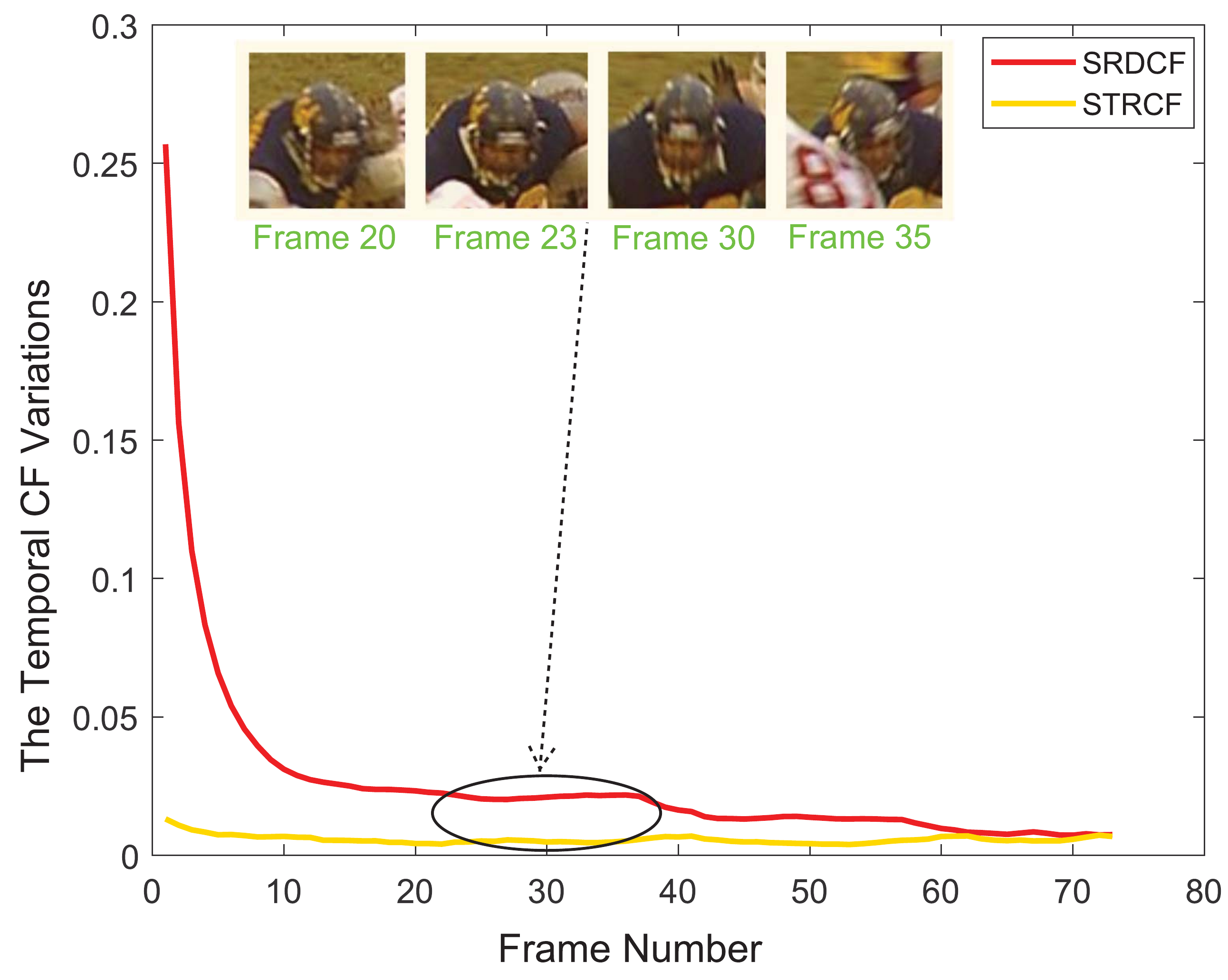}}\
\subfloat[\emph{Jogging}]{\label{fig:Jogging}
  \includegraphics[width=0.37\textwidth,angle=90]{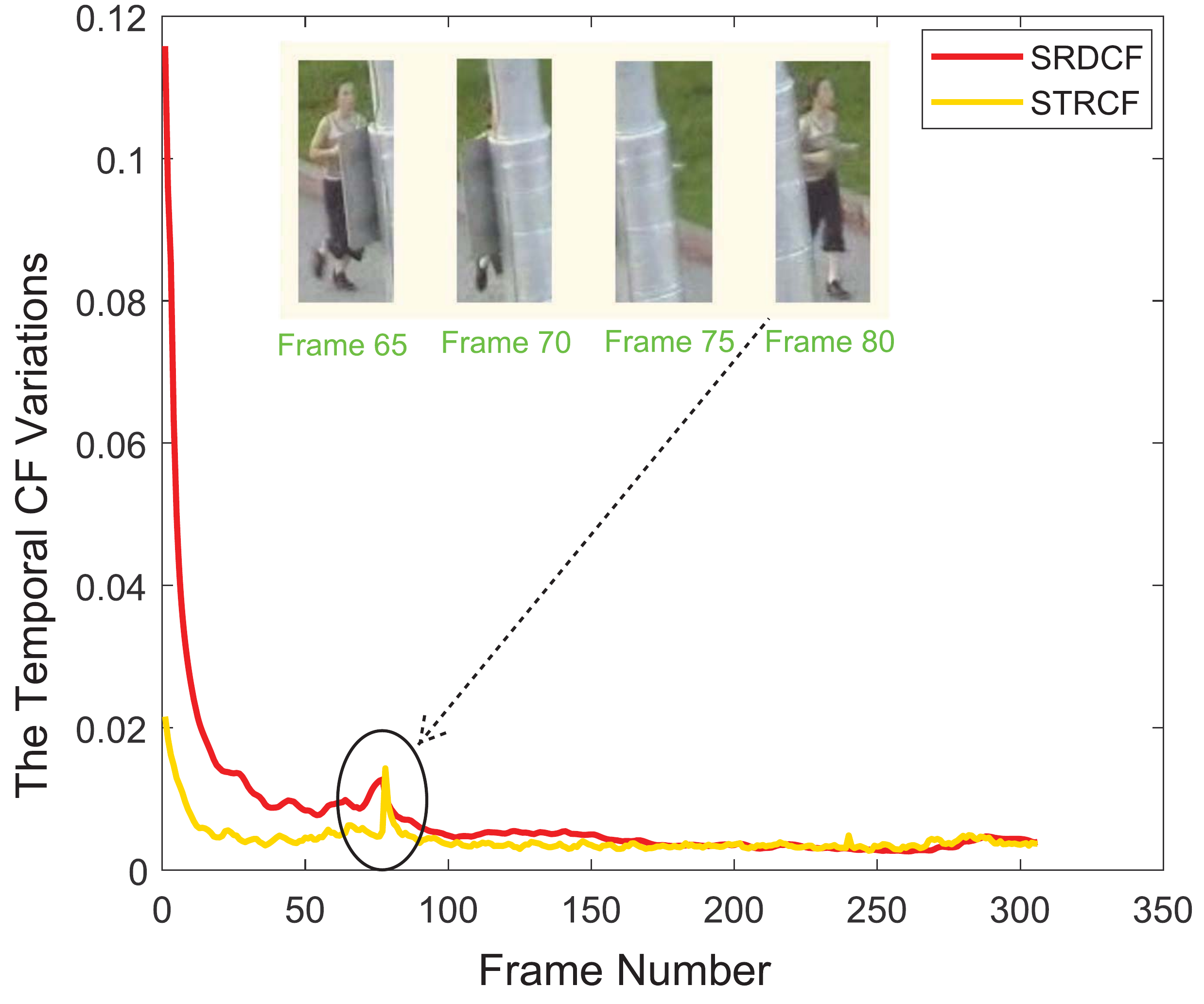}}
\caption{Comparison of the temporal CF variation against frames between SRDCF and our STRCF on 6 video sequences (\ie \emph{Bolt}, \emph{Bolt2}, \emph{Panda},  \emph{Football1}, \emph{DragonBaby} and \emph{Jogging}).}
\label{fig:ALLCFVars}
\end{figure*}
\subsubsection{Effect of regularization parameter $\mu$}

We further analyze the effect of regularization parameter $\mu$ on the tracking performance of {STRCF with hand-crafted features.}
The regularization parameter $\mu$ determines the rate at which to replace the learned CF $\mathbf{f}_{t-1}$ from previous frames with the new sample $\mathbf{x}$ in the current frame.
The lower the parameter $\mu$, the higher relevance of filter $\mathbf{f}$ given to the sample $\mathbf{x}$.
In Fig. \ref{fig:RegularParams}, it is shown that the accuracy of STRCF tracker is significantly affected by the choice of $\mu$.
From Fig. \ref{fig:RegularParams},  the best performance is achieved around $\mu = 16$.
{Note that
when $\mu = 0$, STRCF is trained only with the current frame and
ignores all historical information, thus it even performs worse than KCF.}

\begin{figure}[!htbp]
\setlength{\abovecaptionskip}{-0.cm}
\setlength{\belowcaptionskip}{-0.2cm}
\centering
\subfloat{%
  \includegraphics[width=0.3\textwidth]{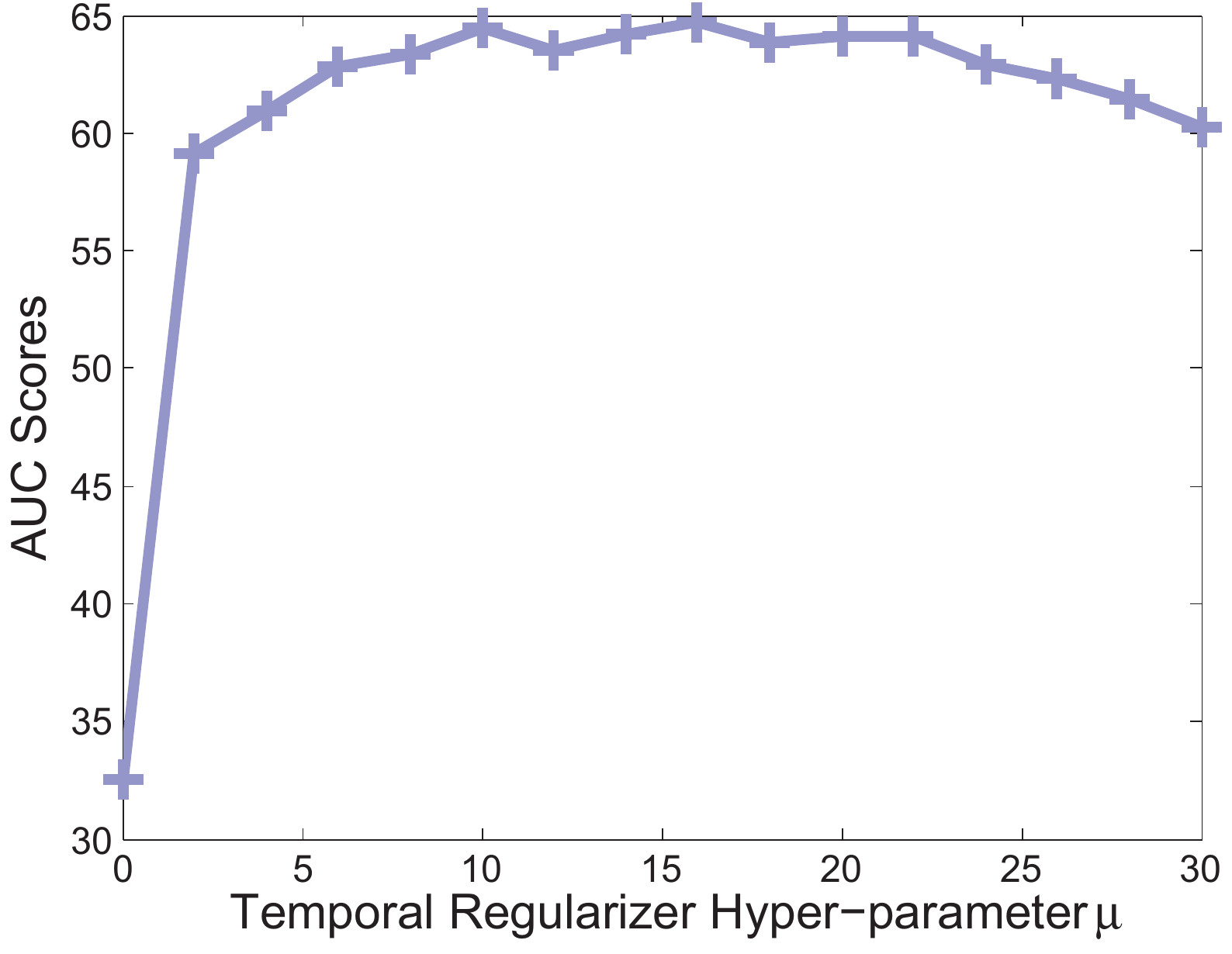}}\\
\caption{\small{Impacts of the temporal regularization parameter $\mu$ on OTB-2015 dataset.}}
\label{fig:RegularParams}
\end{figure}
\begin{figure}[!htbp]
\setlength{\abovecaptionskip}{-0.cm}
\setlength{\belowcaptionskip}{-0.2cm}
\centering
\subfloat{%
  \includegraphics[width=0.3\textwidth]{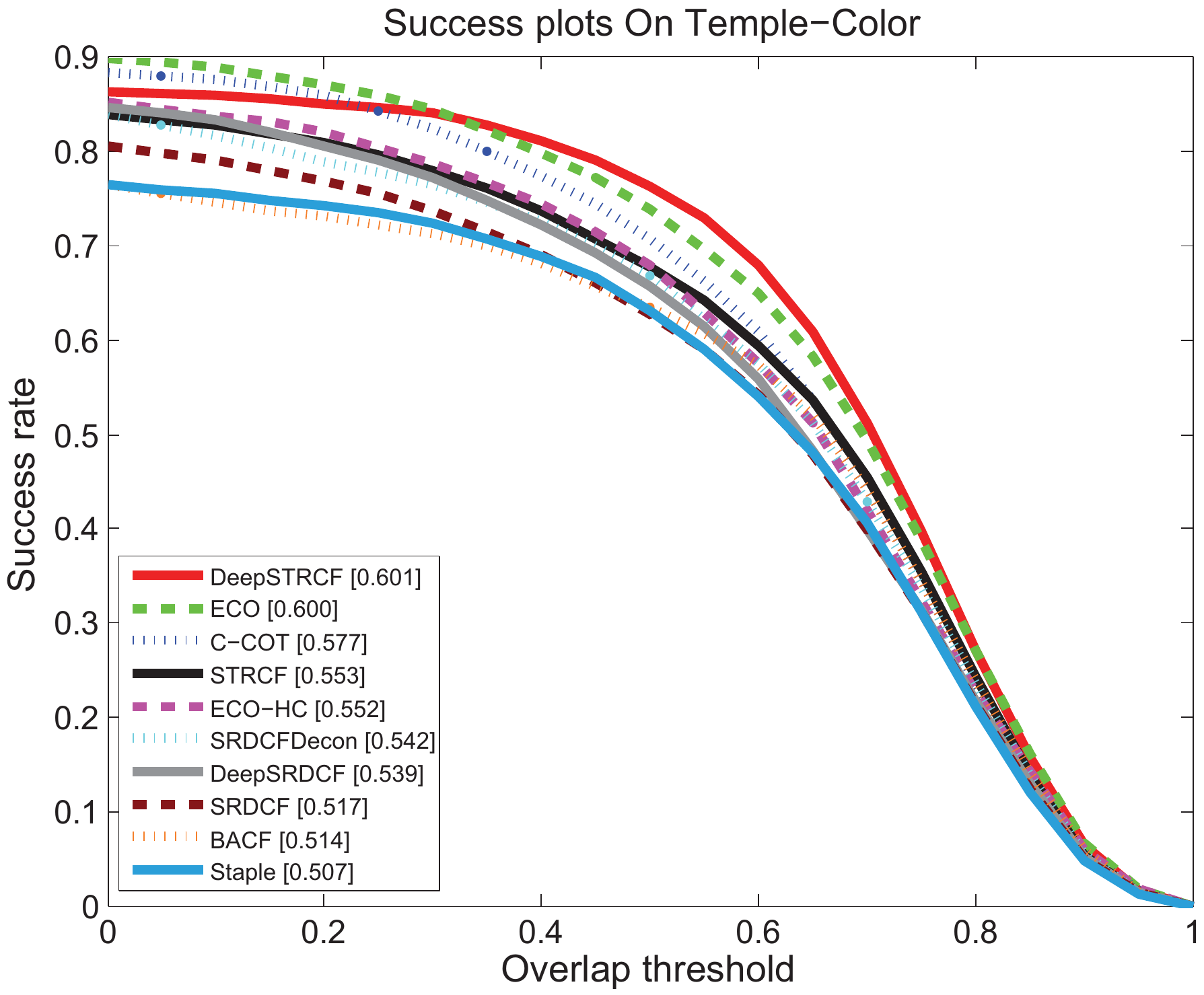}}\\
\caption{\small{The overlap success plot of different trackers on Temple-Color. Only the top 10 trackers are displayed for clarity.}}
\label{fig:Temple}
\end{figure}
\subsection{The Temple-Color Benchmark}

We perform comparative experiments on Temple-Color dataset \cite{Liang2015Encoding} which consists of 128 color sequences.
We compare STRCF and {DeepSTRCF} with the state-of-the-art trackers mentioned above except CF-Net \cite{valmadre2017end} which trained the network on Temple-Color .
Fig. \ref{fig:Temple} shows the comparison of overlap success plots for different trackers.
{We note that STRCF is on par with ECO-HC and surpasses its counterparts SRDCF, DeepSRDCF and SRDCFDecon by 3.6\%, 1.4\% and 1.1\%, respectively.
Meanwhile, DeepSTRCF performs the best among the competing trackers and achieves an AUC score of 60.1\%, further demonstrating the effectiveness of STRCF on deep features.}

\subsection{The VOT-2016 Benchmark}

We also report the results on Visual Object Tracking 2016 benchmark (VOT-2016) \cite{kristan2016visual}, which consists of 60 challenging videos.
We evaluate the trackers in terms of accuracy, robustness and expected average overlap (EAO) \cite{vcehovin2016visual}.
The accuracy measures the average overlap ratio between the predicted bounding box and the ground-truth.
The robustness computes the average number of tracking failures over the sequence.
And EAO averages the no-reset overlap of a tracker on several short-term sequences.

We compare STRCF {and DeepSTRCF} with state-of-the-art trackers,
including MDNet \cite{nam2016mdnet} (VOT-2015 winner) and TCNN \cite{nam2016modeling} (VOT-2016 winner).
Table \ref{tab:VOT2016} lists the results of different trackers on VOT-2016 dataset.
We can see from Table \ref{tab:VOT2016} that STRCF performs significantly better than the BACF and SRDCF methods in terms of the EAO metric.
{In addition, DeepSTRCF also performs favorably against its counterpart DeepSRDCF by a gain of 3.7\% in EAO metric.}

\section{Conclusion}
In this paper, we propose the spatial-temporal regularized correlation filters (STRCF) to address the inefficiency problem of SRDCF.
By introducing the temporal regularizer to SRDCF formulation with single sample, STRCF serves as an approximation of SRDCF with multiple training samples.
Moreover, as an extension of online PA, STRCF can adaptively balance the tradeoff between aggressive and passive model learning, thus leading to more robust models in the case of large appearance variations.
An ADMM algorithm is developed to solve the STRCF model.
We perform experiments on three benchmarks, and the results show that STRCF with hand-crafted features is superior than the baseline SRDCF by accuracy and speed.
Moreover, STRCF with deep features also performs favorably against state-of-the-art trackers in terms of accuracy and robustness.
{In future, we will further improve our STRCF by investigating whether the temporal regularizer can be compatible to SAMF+AT~\cite{Bibi2016Target}, Staple+CA~\cite{cfcatracking}, and the GMM and continuous convolution in ECO~\cite{Danelljan2016ECO}.}


{\small
\bibliographystyle{ieee}
\bibliography{egbib}
}

\end{document}